\documentclass{article}

\usepackage[table]{xcolor} %
\PassOptionsToPackage{sort&compress}{natbib}
\usepackage[preprint]{corl_2026} %

\usepackage{xspace}
\usepackage{multirow}
\usepackage{booktabs}
\usepackage{hhline}
\usepackage[listings]{tcolorbox}
\usepackage{enumitem}
\usepackage{pifont}
\usepackage{dsfont}
\usepackage{xcolor} %
\usepackage{mathrsfs}
\usepackage{diagbox}
\usepackage{caption}
\usepackage{wrapfig}
\usepackage{bbm}
\usepackage{bibunits}

\definecolor{Gray}{gray}{0.85}
\definecolor{White}{gray}{1}
\definecolor{DGray}{gray}{0.8}
\definecolor{WhiteGray}{rgb}{0.9, 0.9, 0.9}
\definecolor{citecolor}{HTML}{0071bc}
\definecolor{darkred}{rgb}{0.6, 0.1, 0.05}
\definecolor{GreenColor}{rgb}{0.137,0.573,0.565}
\definecolor{RedColor}{rgb}{0.949,0.275,0.224}
\definecolor{PurpleColor}{HTML}{8B008B}
\definecolor{OrangeColor}{rgb}{0.914,0.541,0.141}
\definecolor{LightCyan}{rgb}{0.88,1,1}

\definecolor{bestcolor}{rgb}{1, 0.5, 0.25}
\definecolor{secondbestcolor}{rgb}{1, 0.8, 0.5}

\newcommand{\del}[1]{}

\DeclareMathAlphabet\mathbfcal{OMS}{cmsy}{b}{n}

\newcommand{\qheading}[1]{\noindent\mbox{\textbf{#1}\;}}

\newlength\savewidth

\newcommand{\method}{\textbf{\textsc{TeleDexter}}\xspace}

\usepackage{amsmath, amssymb}
\newcommand{\mbold}[1]{\boldsymbol{#1}}
\newcommand{\real}{\mathbb{R}}

\usepackage[capitalize]{cleveref}
\crefname{algorithm}{Alg.}{Algs.}
\Crefname{algocf}{Algorithm}{Algorithms}
\crefname{section}{Sec.}{Secs.}
\Crefname{section}{Section}{Sections}
\crefname{table}{Tab.}{Tabs.}
\Crefname{table}{Table}{Tables}
\crefname{figure}{Fig.}{Figs.}
\Crefname{figure}{Figure}{Figures}
\crefname{equation}{Eq.}{Eqs.}
\Crefname{equation}{Equation}{Equations}
\crefname{appendix}{Appx.}{Appxs.}
\Crefname{appendix}{Appendix}{Appendices}

\makeatletter
\DeclareRobustCommand\onedot{\futurelet\@let@token\@onedot}
\def\@onedot{\ifx\@let@token.\else.\null\fi\xspace}

\makeatother

\usepackage{acronym}
\acrodef{rl}[RL]{Reinforcement Learning}
\acrodef{ddpm}[DDPM]{Denoising Diffusion Probabilistic Model}
\acrodef{ddim}[DDIM]{Denoising Diffusion Implicit Model}
\acrodef{mlps}[MLPs]{Multi-layer Perceptrons}
\acrodef{mdp}[MDP]{Markov Decision Process}
\acrodef{ik}[IK]{inverse kinematics}

\newcommand{\taskname}[1]{\texttt{#1}\xspace}

\newcommand{\taskCylinder}{\taskname{CylinderReorient}}
\newcommand{\taskCuboid}{\taskname{CuboidReorient}}
\newcommand{\taskBunny}{\taskname{BunnyReorient}}
\newcommand{\taskHammer}{\taskname{HammerUse}}
\newcommand{\taskBrush}{\taskname{BrushSweep}}
\newcommand{\taskScrewdriver}{\taskname{ScrewdriverUse}}
\newcommand{\taskBulb}{\taskname{BulbReplace}}
\newcommand{\taskBulbPolicy}{\taskname{BulbInstall}}
\newcommand{\taskHammerPolicy}{\taskname{HammerDriver}}
\newcommand{\taskBrushPolicy}{\taskname{BrushForward}}

\newcommand{\SR}{\ensuremath{\mathbf{SR}}\xspace}
\newcommand{\TP}{\ensuremath{\mathbf{TP}}\xspace}

\title{Towards Human-level Dexterous Teleoperation}

\author{Puhao Li$^{1,2,\ast}$, Zeyuan Chen$^{2,3,\ast}$, Yingying Wu$^{1,2,\ast}$, Pengkun Wei$^{2}$, Yuyang Li$^{2,3}$, Tianyu Wang$^{2,3}$, \\ \textbf{Jiaxiao Shi$^{2}$, Mingrui Yu$^{1}$, Baoxiong Jia$^{2}$, Song-Chun Zhu$^{1,2,3}$, Tengyu Liu$^{2,\dagger}$, Siyuan Huang$^{2,\dagger}$} \\[4pt]
$^1$Tsinghua University \quad $^2$State Key Lab of General Artificial Intelligence, BIGAI \\
$^3$Peking University \quad $^\ast$Equal Contribution \quad $^\dagger$Corresponding author\\[4pt]
{\hypersetup{urlcolor=orange}\url{https://bigai-dex.github.io/blog/teledexter}}\vspace{-8pt}
}

\begin{document}
\newcommand{\teaserCaption}{
    \method learns diverse dexterous in-hand manipulation skills within a single-stage framework, marking a concrete step towards \textbf{human-level dexterous teleoperation}. 
}

\maketitle

\begin{figure}[h]
    \centering
    \vspace{-1em}
    \includegraphics[width=\linewidth]{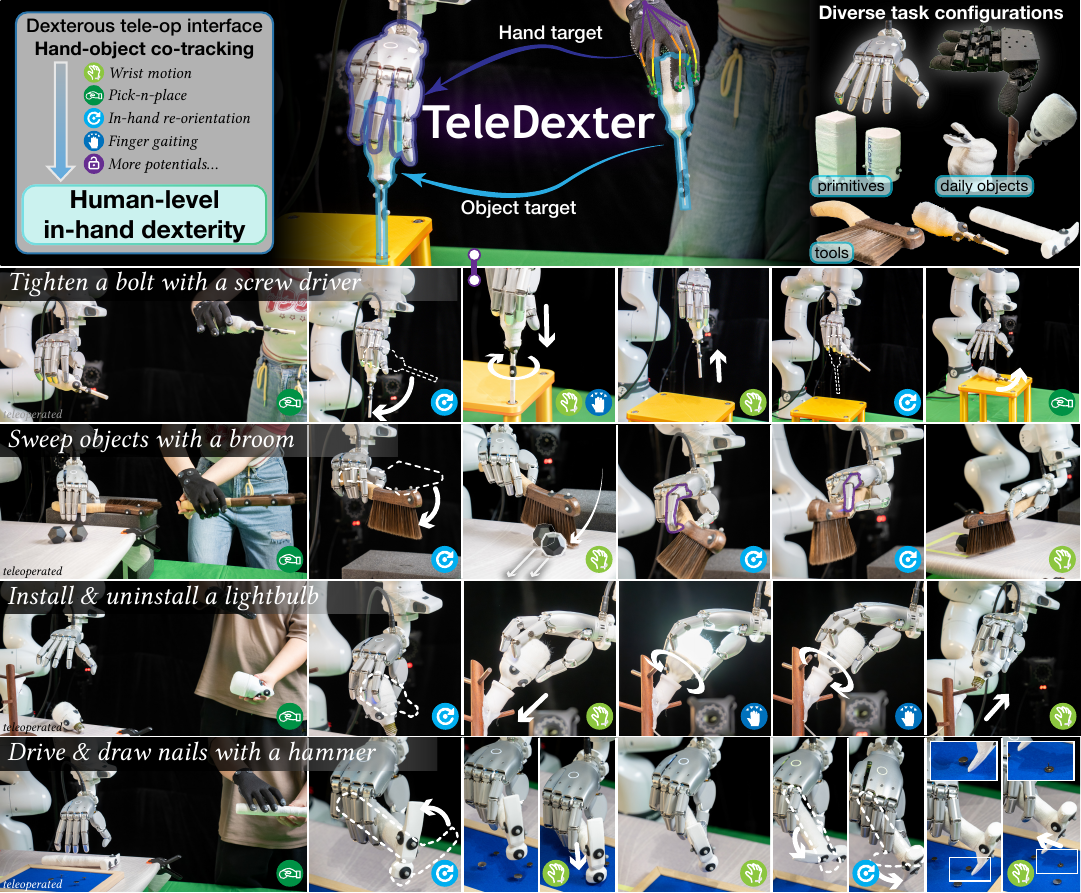}
    \caption{\teaserCaption}
    \label{fig:teaser}
\end{figure}

\begin{abstract}
Humans routinely wield tools, swap grasps, and reposition objects within a single hand—seamlessly orchestrating contact transitions that span translation, re-orientation, and finger gaiting. Endowing robot dexterous hands with this level of in-hand dexterity through teleoperation requires precise control of object motion via dynamic hand–object contact, yet current teleoperation systems remain far from this capability. To bridge this gap, we take a major step towards human-level dexterous teleoperation by introducing \method, a hand-object co-tracking controller that maps operator intent into learned, low-level contact execution. The controller is trained on consecutive co-tracking subgoals derived from human reference motions, utilizing a hybrid reward that couples sparse subgoal objectives with dense tracking rewards to enable learning across diverse interaction modalities rather than frame-wise trajectory imitation. The entire pipeline requires only single-stage RL and, with random action masking and domain randomization, transfers zero-shot to the real robot. We evaluate \method on seven challenging dexterous teleoperation tasks spanning object reorientation and long-horizon tool use across two dexterous hands, achieving a 75\% average success rate where all baselines consistently fail. Furthermore, the collected demonstrations successfully train autonomous policies via behavioral cloning, marking a concrete step towards human-level dexterous teleoperation.
\end{abstract}

\section{Introduction}
\label{sec:intro}

Everyday manipulation demands human-level dexterity: the ability to dynamically reorient, translate, and regrasp objects within a single hand by continuously coordinating complex finger–object contacts~\citep{billard2019trends,bullock2013handcentric}.
While such agility is effortless for humans, it remains far beyond the reach of current robots.
Teleoperation offers a powerful paradigm for closing this gap by enabling operators to directly teach the robot in the loop~\citep{zhang2018deep,mandlekar2020human,qin2023anyteleop}.
\textit{However, achieving human-level in-hand dexterity through dexterous teleoperation remains an open challenge.}

Despite recent progress, existing dexterous teleoperation systems still fall short of human-level in-hand dexterity. One dominant line of work employs kinematic retargeting to map human hand motion directly onto robotic topologies, leveraging vision-based tracking~\citep{handa2020dexpilot,qin2023anyteleop,ding2025bunny,cheng2024open} or wearable exoskeleton gloves~\citep{zhang2025doglove,fang2025dexop,zhu2026dexexo,yin2025geometric}. While this paradigm provides an intuitive, low-latency interface that faithfully captures the operator's kinematic intent, it completely ignores hand-object contact forces and object inertia. Consequently, high-acceleration maneuvers, non-prehensile interactions, and continuous finger-gaiting remain highly unstable, frequently resulting in object slippage or drop failures.

An alternative paradigm~\citep{yin2025dexteritygen} learns a dexterous action prior in simulation to map coarse teleoperation commands to contact-rich hand actions, improving local contact robustness over pure kinematics. However, these methods typically rely on synthetically generated grasp transitions as training targets, which often lack physical feasibility. Furthermore, encoding the action prior into a generative model introduces cascading trajectory drift and covariate shift during closed-loop execution, severely degrading real-world performance during long-horizon deployment.

To overcome these limitations, we introduce \method, \textit{a hand–object co-tracking controller designed for human-level dexterous teleoperation}. Instead of mapping hand joints in isolation, the operator specifies explicit, synchronized geometric targets for both the fingertip positions and the object pose. A low-level controller, trained entirely in simulation with \ac{rl}, then handles the complex multi-contact physics necessary to physically realize these dual co-tracking goals in real time. The key technical designs for \method are three-fold:

\begin{itemize}[leftmargin=*,noitemsep,nolistsep,topsep=0pt,partopsep=0pt]
    \item \textbf{Consecutive subgoal co-tracking:} Rather than forcing rigid, frame-by-frame trajectory imitation, we decompose human reference motions into a sequence of synchronized fingertip and object pose subgoals. By training the policy to reach these consecutive targets rather than blindly copying exact motion configurations, the system gains the operational flexibility needed to discover physically feasible contact-switching strategies. This framework is optimized via a single-stage \ac{rl} pipeline that couples sparse subgoal rewards with dense tracking rewards, eliminating the need for complex, task-specific reward engineering.

    \item \textbf{Co-tracking sequence construction:} We introduce a geometry-aware retargeting pipeline that translates unscripted human hand-object demonstrations into physically grounded reference motions. Going beyond pure kinematic mapping, our approach utilizes a two-stage optimization that incorporates object meshes to enforce contact-surface attraction and penalize mesh interpenetration. This process yields synchronized sequences of fingertip targets and object poses that serve as geometrically feasible subgoals for the low-level controller.
    
    \item \textbf{Sim-to-real robustness: } To ensure robust zero-shot sim-to-real transfer, we introduce random action masking as a strong action-space regularizer. This technique prevents the policy from overfitting to perfectly synchronized simulated actuation, which, when combined with systematic domain randomization, allows the controller to deploy directly onto real robots.
\end{itemize}

We evaluate \method on seven challenging dexterous teleoperation tasks across two distinct dexterous hand embodiments. These tasks range from object rearrangement requiring continuous in-hand reorientation to long-horizon tool use with a hammer, screwdriver, brush, and light bulb. \textbf{\method achieves an average success rate of 75\% across these tasks, whereas baseline models consistently fail.} Furthermore, we demonstrate that the high-quality teleoperation demonstrations collected via \method can be directly leveraged to train fully autonomous policies via behavioral cloning, achieving closed-loop manipulation capabilities without a human in the loop during long-term and dexterous task execution. Ablation studies confirm that our consecutive subgoal tracking and reward design are essential for learning diverse in-hand manipulation modalities in a single stage, while random action masking successfully bridges the sim-to-real gap. Broadly, this work provides a scalable foundation for collecting rich, in-hand dexterous manipulation data, unlocking a viable path toward human-level robotic dexterity.

\section{Related Work}
\label{sec:related}

\qheading{Dexterous Teleoperation}
provides a powerful paradigm for transferring human dexterity to robotic hands~\citep{niemeyer2016telerobotics,zhang2018deep,hedayati2018improving,du2012markerless,kofman2007robot}.
One dominant line of work is kinematic retargeting, which kinematically translates the human hand configuration to robot joint positions, via vision-based tracking \citep{handa2020dexpilot, qin2023anyteleop, ding2025bunny, cheng2024open, yang2024ace}, wearable or exoskeleton gloves \citep{wang2024dexcap, zhang2025doglove, fang2025dexop, zhu2026dexexo}, or learned neural mappings \citep{yin2025geometric}. This paradigm provides an intuitive, low-latency interface but lacks a dynamics prior, making contact-rich actions such as in-hand reorientation, finger gaiting, and tool use infeasible. To address this limitation, DexGen \citep{yin2025dexteritygen} learns a generative dexterous action prior from simulation rollouts that maps coarse teleoperation commands to fine hand actions, improving contact robustness. However, its reliance on synthetically generated grasp transitions as training goals does not guarantee physical feasibility, and encoding the action prior into a generative model introduces compounding errors that degrade real-world performance. In contrast, we directly train an \ac{rl} controller guided from human hand–object reference motions, providing physically grounded goals that cover diverse in-hand manipulation modalities. The learned controller directly serves as the robust but agile low-level teleoperation policy, enabling human-level in-hand dexterity including long-horizon tool use.

\qheading{Learning Dexterous Manipulation via \ac{rl}}
in simulation has driven rapid progress, from grasping~\cite{li2023gendexgrasp,xu2023unidexgrasp,li2024grasp,li2025maniptrans}, in-hand reorientation \citep{andrychowicz2020learning, akkaya2019solving, handa2023dextreme, qi2023inhand,chen2023visual, yang2024anyrotate,liu2025dexndm} to complex finger gaiting and dynamic skills \citep{qi2025simple,wang2024lessons}. However, these approaches typically learn task-specific policies with dedicated reward engineering for each skill. Recent work mitigates this by using human hand–object interaction data as reference motions to guide RL, enabling diverse manipulation skills without per-task reward design \citep{liu2024parameterized, chen2024object, li2025maniptrans, liu2025dextrack}. However, these methods primarily learn one policy per trajectory and rarely achieve in-hand dexterous skills such as reorientation or finger gaiting. We attribute this to the dense frame-wise tracking formulation, which is overly restrictive for single-stage policy learning and prevents the RL agent from exploring dynamic contact strategies beyond basic grasping and wrist motion. To address this, we introduce a consecutive subgoal tracking formulation that learns from human hand–object reference motions, enabling diverse and dynamic in-hand skills within a single RL training stage. Combined with the proposed random action masking and systematic domain randomization \citep{peng2018sim, chen2023visual}, the learned policy transfers zero-shot to the real robot as a dexterous teleoperation controller.

\section{\method}
\label{sec:method}
\begin{figure*}[!t]
    \centering
    \includegraphics[width=\linewidth]{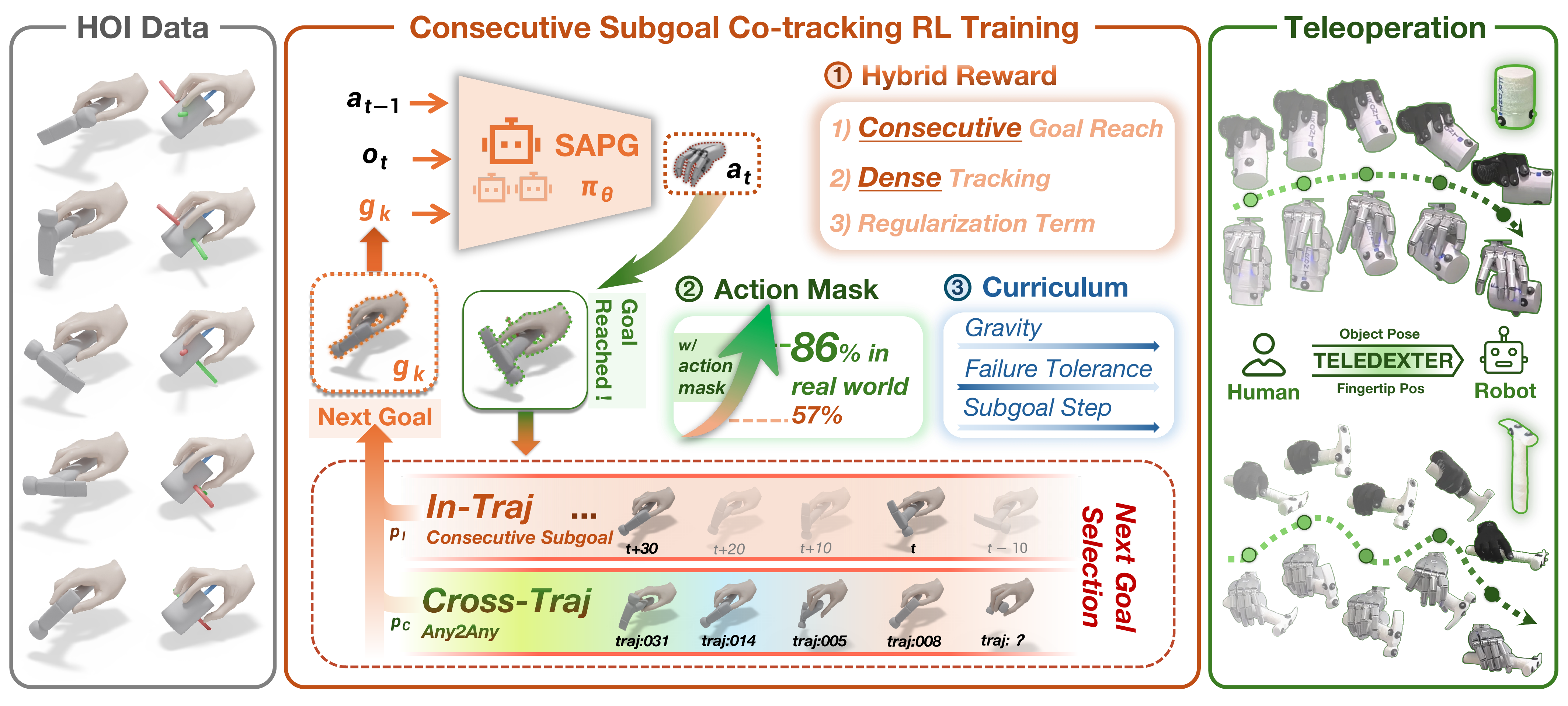}
    \caption{\textbf{Method overview of \method.}
    Given human hand-object reference motions, we train a co-tracking controller via single-stage RL and deploy it zero-shot to real-world dexterous teleoperation.
    }
    \label{fig:method_overview}
    \vspace{-12pt}
\end{figure*}

We formulate dexterous teleoperation as hand–object co-tracking. The operator specifies hand and object pose targets, and our learned low-level controller executes the multi-contact dynamics to physically reach these goals. As shown in \cref{fig:method_overview}, we first formulate the co-tracking problem given a set of hand-object reference motions, then present the single-stage \ac{rl} framework for training the co-tracking controller (\cref{sec:method:learning}). We then describe how these reference motions are constructed (\cref{sec:method:refmotion}) and deploy the learned policy to the real world as the teleoperation controller (\cref{sec:method:deployment}).

\qheading{Problem Formulation}
All quantities below are expressed in the wrist frame. The robot arm tracks the human wrist pose independently via \ac{ik}. Given a set of hand-object reference motions (\cref{sec:method:refmotion}), our goal is to learn a co-tracking policy $\mbold{a}_t = \pi_{\theta}(\mbold{o}_t,\, g_t)$ that drives the robot hand and the manipulated object toward target poses prescribed by a co-tracking goal $g_t$. Here $\mbold{o}_t$ encodes the current robot hand-object state, $\mbold{a}_t \in \real^{n_{\mathrm{dof}}}$ are target joint positions, and the co-tracking goal specifies both target fingertip positions and target object pose:
\(
    g_t = \bigl(\hat{\mbold{p}}^{\mathrm{tip}}_t,\; \hat{T}^{o}_t\bigr)
\),
where
\(
    \hat{\mbold{p}}^{\mathrm{tip}}_t \in \real^{N_f \times 3}
\)
and
\(
    \hat{T}^{o}_t = \bigl(\hat{\mbold{x}}^{o}_t,\; \hat{R}^{o}_t\bigr) \in SE(3)
\).
During teleoperation, $g_t$ is constructed from real-time captured hand-object poses, casting dexterous teleoperation as a co-tracking problem. This formulation prescribes \emph{what} the hand and object should achieve, while leaving the contact strategy, i.e.\@ \emph{how}, to the learned controller.

\subsection{Learning a Co-tracking Controller}
\label{sec:method:learning}
Given the reference motions for the manipulated object, we train a hand-object co-tracking controller $\pi_\theta$ in simulation via \ac{rl}.
The co-tracking goals guide the controller to learn a dynamic hand-object action prior through simulated contact, moving beyond kinematic trajectory imitation.
The controller input consists of the observation $\mbold{o}_t$ and the co-tracking goal $g_t$.
The observation $\mbold{o}_t$ contains the current hand joint positions $\mbold{q}_t$, the object pose $({\mbold{x}}^{o}_t, {R}^{o}_t)$ and the gravity direction in the wrist frame, and the previous action $\mbold{a}_{t-1}$.

\qheading{Consecutive Subgoal Co-tracking}
Each reference trajectory is converted into a sequence of co-tracking 
subgoals sampled at varying intervals. We term this formulation 
\emph{consecutive subgoal co-tracking}: the policy must reach each 
hand-object subgoal in order before advancing to the next, while freely 
discovering its own contact strategy between subgoals.
For a subgoal $g_k = (\hat{\mbold{p}}^{\mathrm{tip}}_k, \hat{T}^{o}_k)$, the per-finger, object position, and object rotation tracking errors are
\begin{equation}
    e^{f}_{t,k} = \left\|\mbold{p}^{\mathrm{tip}}_{t,f} - \hat{\mbold{p}}^{\mathrm{tip}}_{k,f}\right\|_2,\quad
    e^{\mathrm{pos}}_{t,k} = \left\|\mbold{x}^{o}_t - \hat{\mbold{x}}^{o}_k\right\|_2,\quad
    e^{\mathrm{rot}}_{t,k} = \left\|\mathrm{Log}\!\left(\hat{R}^{o\top}_k R^o_t\right)\right\|_2.
    \label{eq:tracking_errors}
\end{equation}
A subgoal is reached when, for \(N_{\mathrm{stay}}\) consecutive frames, 
\(\max_{f} e^{f}_{t,k} < \epsilon_{\mathrm{tip}}\),
\(e^{\mathrm{pos}}_{t,k} < \epsilon_{\mathrm{pos}}\), and
\(e^{\mathrm{rot}}_{t,k} < \epsilon_{\mathrm{rot}}\).
Once this criterion is satisfied, the policy advances to the next subgoal.

\qheading{Hybrid Reward Design}
The dominant learning signal is consecutive goal reaching, augmented with 
dense tracking reward for early exploration.
The reward function combines a \emph{consecutive goal-reaching reward} 
with \emph{dense tracking reward}:
\begin{equation}
r_t \;=\; \underbrace{\mathbbm{1}_{\mathrm{reach}}(t)\,w_{\mathrm{step}}(t)\,r_{\mathrm{score}}(t)}_{\text{sparse subgoal}}
\;+\; \underbrace{\alpha_{\mathrm{dense}}\,r_{\mathrm{dense}}(t)}_{\text{dense tracking}}
\;-\; \underbrace{c_{\mathrm{time}}}_{\text{time}},
\end{equation}
The indicator $\mathbbm{1}_{\mathrm{reach}}(t)$ fires when the active subgoal is reached, and $w_{\mathrm{step}}$ weights the reward by the inter-subgoal step size. The score $r_{\mathrm{score}}$ measures how well the hand and object match the active subgoal:
\begin{equation}
    r_{\mathrm{score}}
    =
    \sum_{f=1}^{N_f} w_{f}\exp\!\left(-\beta_{f}\, e^{f}_{t,k}\right)
    +
    w_{\mathrm{pos}}\exp\!\left(-\beta_{\mathrm{pos}}\, 
    e^{\mathrm{pos}}_{t,k}\right)
    +
    w_{\mathrm{rot}}\exp\!\left(-\beta_{\mathrm{rot}}\, 
    e^{\mathrm{rot}}_{t,k}\right).
\end{equation}
The dense tracking reward $r_{\mathrm{dense}}$ uses the same per-finger and 
object tracking terms at every timestep, scaled by $\alpha_{\mathrm{dense}}$, 
providing a small dense signal during early training.

\qheading{Curriculum Learning}
We progressively increase three dimensions of difficulty during training.
(i)~Gravity is reduced initially and annealed to full gravity, easing 
initial contact establishment.
(ii)~The subgoal tracking tolerances 
$(\epsilon_{\mathrm{tip}}, \epsilon_{\mathrm{pos}}, 
\epsilon_{\mathrm{rot}})$ start permissive 
and are progressively tightened, enforcing stricter tracking precision 
over training.
(iii)~The inter-subgoal step size grows from small to large, so the policy 
first masters fine-grained local tracking and later handles longer-horizon 
goal jumps.
Episodes are initialized at random frames across the reference motions, 
and successful traversal of one trajectory resets the environment to 
another (cross-trajectory reset), enabling continuous learning across the 
full reference-motion set.

\qheading{Sim-to-Real Robustness}
For sim-to-real transfer, we apply domain randomization to tolerate dynamics, sensing, and actuation mismatch between simulation and the real world.
Following~\citep{chen2023visual}, we randomize the shape and dynamics 
properties of both the object and the dexterous hand, apply random external 
forces to the object, and inject observation noise and latency.
In addition, we introduce \emph{random action masking} as a strong 
action-space regularization.
Given the policy action $\mbold{a}_t$, we sample a binary mask 
$\mbold{m}_t \in \{0,1\}^{n_{\mathrm{dof}}}$ and apply 
$\tilde{\mbold{a}}_t = \mbold{m}_t \odot \mbold{a}_t + 
(1-\mbold{m}_t)\odot \tilde{\mbold{a}}_{t-1}$, 
where masked dimensions are frozen at the previous command for a randomly 
sampled duration.
By forcing the policy to succeed even when subsets of joints retain stale 
commands, this regularization prevents the policy from overfitting to 
simulation dynamics, which inevitably differ from those of the real world.

\qheading{Single-stage RL Training}
Each controller is trained in a single RL stage using large-scale parallel simulation, without staged skill decomposition or task-specific reward engineering.
We use Isaac Gym~\citep{makoviychuk2021isaac}, and all reference motions for one object (${\sim}$50 minutes in our setting) are loaded simultaneously.
We use SAPG~\citep{singla2024sapg} to optimize the policy with 4 NVIDIA RTX 5090 GPUs, running ${\sim}$62{,}000 parallel environments.
Training converges within ${\sim}10^{10}$ environment steps.
Being reference-driven, the framework scales directly to new objects, hands and more interaction patterns.

\subsection{Hand-Object Reference Motion Construction}
\label{sec:method:refmotion}
We now describe how the hand-object reference motions used in training are constructed. Starting from human hand-object interaction trajectories recorded by a motion-capture system, we convert them into robot hand-object reference motions through geometry-aware retargeting. The recorded interactions span three categories: \emph{(i)~in-hand translation}, \emph{(ii)~in-hand rotation}, and \emph{(iii)~free-play} combining arbitrary grasps, finger gaiting, and tool-use motion sequences.

\qheading{Geometry-aware Retargeting}
The retargeting proceeds in two stages.
The first stage follows vector-based retargeting~\citep{handa2020dexpilot, qin2023anyteleop}, optimizing robot joint angles $\mbold{q}_{1:T}$ to match human hand geometry via directional and inter-finger vector alignment.
Kinematic retargeting alone does not account for object geometry or contact feasibility.
The second stage refines the result with a geometry-aware optimization that 
incorporates the object mesh.
The final trajectory is obtained by
\begin{equation}
    \mbold{q}^{*}_{1:T}
    =
    \arg\min_{\mbold{q}_{1:T}}
    \sum_{t=1}^{T}
    \left(
    \mathcal{L}_{\mathrm{vec}}^t
    + \lambda_{\mathrm{surf}}\mathcal{L}_{\mathrm{surf}}^t
    + \lambda_{\mathrm{pen}}\mathcal{L}_{\mathrm{pen}}^t
    + \lambda_{\mathrm{col}}\mathcal{L}_{\mathrm{col}}^t
    \right)
    + \lambda_{\mathrm{smooth}}\mathcal{L}_{\mathrm{smooth}},
    \label{eq:retarget}
\end{equation}
where $\mathcal{L}_{\mathrm{vec}}^t$ is the vector-retargeting loss~\citep{qin2023anyteleop, handa2020dexpilot}. Let $\mathcal{H}_t$ and $\mathcal{O}_t$ denote the hand and object meshes at frame $t$, with $\mathrm{sdf}_{\mathcal{O}_t}(\cdot)$ the differentiable signed distance~\citep{kaolin}. The surface term $\mathcal{L}_{\mathrm{surf}}^t = \frac{1}{|\mathcal{S}_t|}\sum_{p \in \mathcal{S}_t} \mathrm{ReLU}(\mathrm{sdf}_{\mathcal{O}_t}(p))$ pulls near-contact hand points ($\mathcal{S}_t = \{p : \mathrm{sdf}_{\mathcal{O}_t}(p) < \tau_{\mathrm{surf}}\}$) onto the object surface. The penetration term $\mathcal{L}_{\mathrm{pen}}^t = \sum_{p \in \mathcal{H}_t} \mathrm{ReLU}\!\bigl(-\mathrm{sdf}_{\mathcal{O}_t}(p)\bigr)$ penalizes interpenetration. The self-collision term $\mathcal{L}_{\mathrm{col}}^t = \frac{1}{2}\sum_{f(i)\ne f(j)} \mathrm{ReLU}\!\bigl((r_i + r_j) - \|\mbold{c}_i - \mbold{c}_j\|_2\bigr)$ prevents inter-finger overlap via collision spheres with radii $r_i, r_j$ and centers $\mbold{c}_i, \mbold{c}_j$. $\mathcal{L}_{\mathrm{smooth}}$ applies the Curobo~\citep{curobo_v2} temporal smoothness energy on $\mbold{q}_{1:T}$ to suppress capture jitter. 
The output is a set of robot hand-object reference trajectories $\{(\mbold{q}^{*}_t, T^o_t)\}_{t=1}^{T}$ in the wrist frame, from which the co-tracking goals used in \cref{sec:method:learning} are constructed as fingertip targets $\hat{\mbold{p}}^{\mathrm{tip}}_t = \mathrm{FK}_{\mathrm{tip}}(\mbold{q}^{*}_t)$ and object targets $\hat{T}^o_t = T^o_t$.

\subsection{Real-world Teleoperation Deployment}
\label{sec:method:deployment}
The learned co-tracking controller deploys zero-shot as the teleoperation controller.
A real-time system captures the operator's wrist, fingertip, and object poses; the arm tracks the wrist via \ac{ik}, while the fingertip and object poses form the co-tracking goal for the policy.
For contact initialization, kinematic retargeting~\citep{handa2020dexpilot} handles pre-grasp positioning; once stable contact is established, the operator switches to the co-tracking controller for dexterous manipulation.

\section{Experiments}
\label{sec:exp}
We conduct a systematic real-world evaluation of \method for dexterous hand teleoperation. In experiments, we describe the experimental setup (\cref{sec:exp:setup}), compare \method against representative baselines on seven dexterous tasks across two hand embodiments (\cref{sec:exp:task}), demonstrate autonomous policy learning from collected teleoperation data (\cref{sec:exp:auto}), and conduct ablation studies on consecutive subgoal tracking and random action masking (\cref{sec:exp:ablation}).

\subsection{Experimental Setup}
\label{sec:exp:setup}
\qheading{Robot Platforms}
All real-world experiments are conducted on a Franka FR3 arm equipped with a dexterous robot hand. We evaluate two hand embodiments: LeapHand \citep{shaw2023leap}, a four-finger hand with 16 DoFs, and SharpaWave, a five-finger human-like hand with 22 DoFs, to validate that our framework can generalize across different robotic hand morphologies and actuation spaces.

\qheading{Teleoperation Interface}
We use a NOKOV MoCap system to track the operator's hand pose and the manipulated object's 6D pose in real time. For all methods, the operator's wrist pose is converted into Franka arm commands via \ac{ik}. The  methods differ in how they map the captured hand and object references to dexterous hand actions. All teleoperation evaluations run at 30\,Hz.

\qheading{Baselines}
\textit{DexRT}~\citep{handa2020dexpilot,qin2023anyteleop} directly maps the operator's hand motion to the robot hand through kinematic retargeting. 
\textit{GeoRT}~\citep{yin2025geometric} learns a neural retargeting function using geometric objectives. 
\textit{DexGen}~\citep{yin2025dexteritygen} uses a learned generative action prior to convert teleoperation commands into contact-rich hand actions. 
\textit{SimToolReal}~\citep{kedia2026simtoolreal} is an object-centric sim-to-real tool manipulation method that requires human reference motions at inference time to guide execution. Although not a teleoperation approach, it provides a strong reference for learned dexterous tool use.

\subsection{Dexterous Teleoperation Evaluation}
\label{sec:exp:task}

\begin{figure}[!ht]
    \vspace{-4pt}
    \centering
    \includegraphics[width=\linewidth]{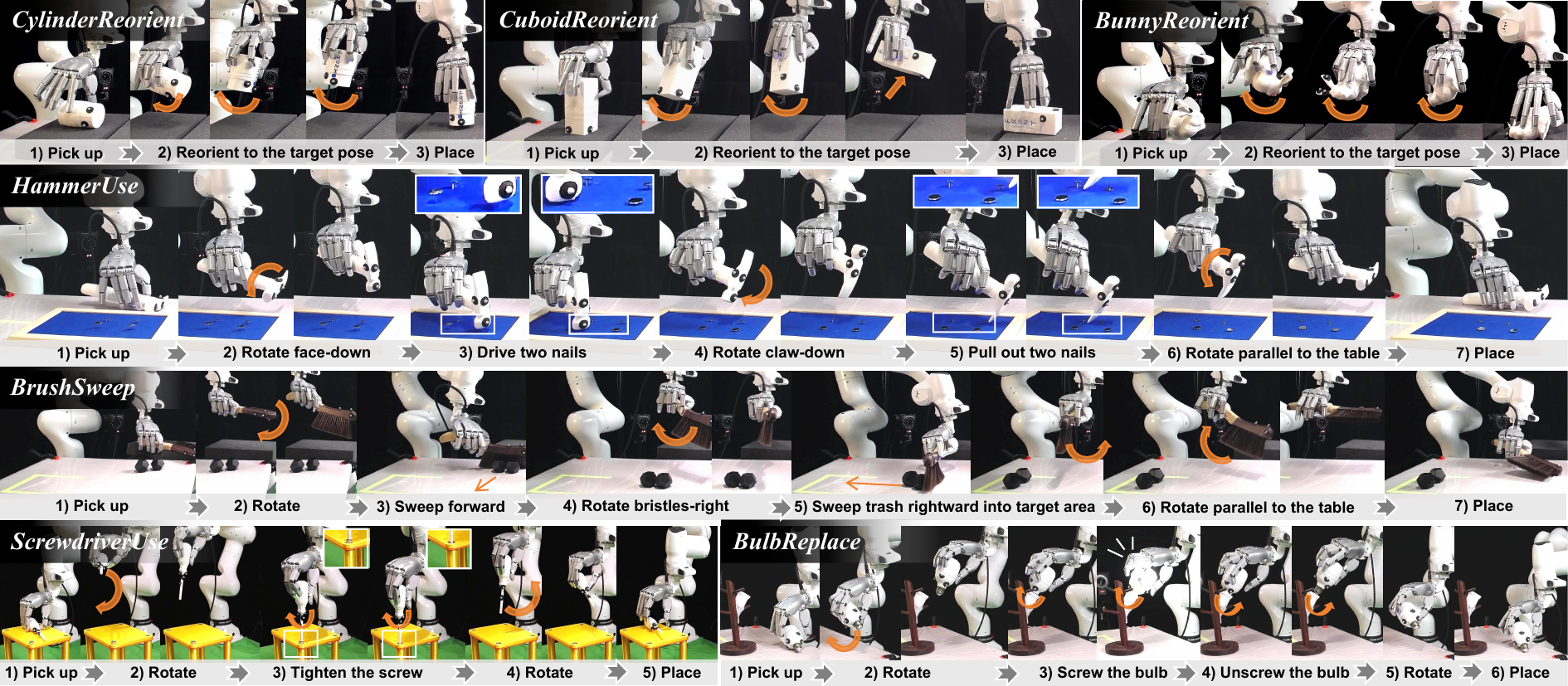}
    \caption{\textbf{Task descriptions.}
    Seven dexterous tasks across two categories: three reorientation tasks over diverse geometries and four long-horizon tool-use tasks. Each task is decomposed into well-defined stages.
    }
    \label{fig:task_description}
    \vspace{-6pt}
\end{figure}

\qheading{Overview}
As shown in \cref{fig:task_description}, we evaluate \method on seven real-world dexterous tasks spanning two categories. The three reorientation tasks (\taskCylinder, \taskCuboid, \taskBunny) test precise in-hand pose control over symmetric, edge/corner, and irregular geometries. 
The four tool-use tasks (\taskHammer, \taskBrush, \taskScrewdriver, \taskBulb) require long-horizon multi-stage manipulation involving in-hand reorientation to transition between functional grasps, finger gaiting, and tool application. 
Together, these tasks provide a comprehensive evaluation of whether a teleoperation system can robustly execute diverse dexterous manipulation.

\begin{table*}[!t]
    \centering
    \captionsetup{font=small}
    \caption{\textbf{Dexterous teleoperation results on SharpaWave.}
        Each cell: \SR\,/\,\TP (\%; higher is better). \textit{SimToolReal} is not a teleoperation method ($\dagger$ = category-specific, $\ddagger$ = all categories) averaged over three tasks only.
        }
    \label{tab:task_results}
    \small
    \setlength{\tabcolsep}{4pt}
    \resizebox{\linewidth}{!}{%
        \begin{tabular}{lcccccc}
            \toprule
            \textbf{Task}
            & \textit{DexRT}
            & \textit{GeoRT}
            & \textit{DexGen}
            & \textit{SimToolReal}$^{\dagger}$
            & \textit{SimToolReal}$^{\ddagger}$
            & \method \\
            \midrule
            \taskCylinder
            &  6.7 / 37.8  &  0.0 / 24.4  &  0.0 / 31.1
            & \multicolumn{1}{c}{---}  & \multicolumn{1}{c}{---}
            & \textbf{80.0} / \textbf{86.7} \\
            \taskCuboid
            & 26.7 / 51.1  &  0.0 / 33.3  &  0.0 / 28.9
            & \multicolumn{1}{c}{---}  & \multicolumn{1}{c}{---}
            & \textbf{80.0} / \textbf{86.7} \\
            \taskBunny
            &  0.0 / 35.6  &  0.0 / 31.1  &  0.0 / 26.7
            & \multicolumn{1}{c}{---}  & \multicolumn{1}{c}{---}
            & \textbf{66.7} / \textbf{77.8} \\
            \midrule
            \taskHammer
            &  0.0 / 26.7  &  0.0 / 30.5  &  0.0 / 26.7
            &  0.0 / 27.6  & 20.0 / 36.2
            & \textbf{66.7} / \textbf{86.7} \\
            \taskBrush
            &  0.0 / 39.0  &  0.0 / 29.5  &  0.0 /  8.6
            & 26.7 / 41.9  &  0.0 /  5.7
            & \textbf{73.3} / \textbf{89.5} \\
            \taskScrewdriver
            &  6.7 / 37.3  &  0.0 / 25.3  &  0.0 / 33.3
            &  0.0 / 17.3  &  0.0 / 20.0
            & \textbf{73.3} / \textbf{86.7} \\
            \taskBulb
            &  0.0 / 35.6  &  0.0 / 25.6  &  0.0 / 20.0
            & \multicolumn{1}{c}{---}  & \multicolumn{1}{c}{---}
            & \textbf{86.7} / \textbf{95.6} \\
            \midrule
            \textbf{Average}
            &  5.7 / 37.6  &  0.0 / 28.5  &  0.0 / 25.0
            &  8.9 / 28.9  &  6.7 / 20.6
            & \textbf{75.2} / \textbf{87.1} \\
            \bottomrule
        \end{tabular}
    }
    \vspace{-12pt}
\end{table*}
\begin{wraptable}{rt}{0.38\linewidth}
    \centering
    \captionsetup{font=small}
    \caption{\textbf{Teleoperation on LeapHand.}}
    \label{tab:task_results_leap}
    \vspace{-6pt}
    \footnotesize
    \setlength{\tabcolsep}{3pt}
    \begin{tabular}{lc}
        \toprule
        \textbf{Task} & \method \\
        \midrule
        \taskCylinder & 60.0 / 73.3 \\
        \taskCuboid   & 73.3 / 82.2 \\
        \bottomrule
    \end{tabular}
    \vspace{-12pt}
\end{wraptable}

\qheading{Protocol and Metrics}
For each task, we conduct 15 trials per method. Each task is decomposed into a sequence of well-defined stages (shown in \cref{fig:task_description}). We report two metrics. Success rate (\SR) is the percentage of trials completing all stages. Task progress (\TP) is the average percentage of stages completed per trial, where each stage contributes equally. A trial is terminated when an unrecoverable grasp loss or object drop occurs during execution.

\qheading{Results and Analysis}
As shown in \cref{tab:task_results}, \method achieves 75.2\% average \SR and 87.1\% average \TP across all seven tasks, while all baselines near-uniformly fail. On the reorientation tasks, \method achieves 66.7--80.0\% \SR by executing dynamic in-hand reorientation through learned contact strategies. In contrast, kinematic retargeting methods (\textit{DexRT}, \textit{GeoRT}) rarely progress beyond the initial pick-up stage, as they lack the dynamics prior needed for contact-rich in-hand manipulation. \textit{DexGen}, despite its learned action prior, similarly fails due to compounding errors in its generative model that degrade real-world contact execution. The gap widens on tool-use tasks, which demand long-horizon coordination of functional grasp transitions, finger gaiting, and tool application. \method achieves 66.7--86.7\% \SR on these tasks, while all teleoperation baselines achieve near-zero \SR. 
The \TP metric reveals where failures occur: \method's narrow \SR-to-\TP gap (75.2\% vs.\ 87.1\%) indicates that most failures happen at late task stages, whereas baselines consistently collapse at the first stage requiring in-hand reorientation or finger gaiting. On \taskBulb, \method completes both the screw-in and unscrew rotation stages without losing a single trial ($15/15$ through stage~4 of~6), with the only failures at final placement ($13/15$). Similarly, on \taskScrewdriver, $13$ of $15$ trials sustain continuous finger gaiting through the tightening stage, a contact mode that no baseline can execute.
\begin{table}[!t]
    \centering
    \captionsetup{font=small}
    \caption{\textbf{Autonomous policy stage-wise success.}
    \taskBulbPolicy: pick up $\to$ reorient $\to$ align $\to$ install;
    \taskHammerPolicy: pick up $\to$ rotate $\to$ drive nails;
    \taskBrushPolicy: pick up $\to$ rotate $\to$ sweep forward.}
    \label{tab:auto_policy}
    \vspace{4pt}
    \setlength{\tabcolsep}{6pt}
    \begin{tabular}{lccccc}
    \toprule
    \textbf{Task} & Stage 1 & Stage 2 & Stage 3 & Stage 4 & \SR \\
    \midrule
    \taskBulbPolicy   & 13/15 & 12/13 & 8/12 & 7/8 & 46.7\% \\
    \taskHammerPolicy & 15/15 & 15/15 & 11/15 & ---   & 73.3\% \\
    \taskBrushPolicy  & 7/15  & 7/7   & 6/7  & ---   & 40.0\% \\
    \bottomrule
    \end{tabular}
    \vspace{-8pt}
\end{table}
\textit{SimToolReal}, an object-centric policy trained specifically for tool manipulation, achieves at most 26.7\% \SR on a single task and fails on the others, indicating that even dedicated tool-use policies struggle to generalize across the diverse contact transitions required for long-horizon manipulation. 
As shown in \cref{tab:task_results_leap}, applying the same training pipeline to LeapHand yields a strong controller with minimal embodiment-specific tuning. Notably, both controllers are trained from the same human hand-object interaction reference motions; only the geometry-aware retargeting stage adapts to the target morphology (4-finger, 16-DoF LeapHand vs.\ 5-finger, 22-DoF SharpaWave). Despite this substantial morphological gap, LeapHand achieves $60.0$--$73.3\%$ \SR on the reorientation tasks, confirming that the framework generalizes across embodiments without re-collecting human 
\begin{wrapfigure}[18]{r}{0.5\columnwidth}
  \vspace{24pt}
  \centering
  \includegraphics[width=\linewidth]{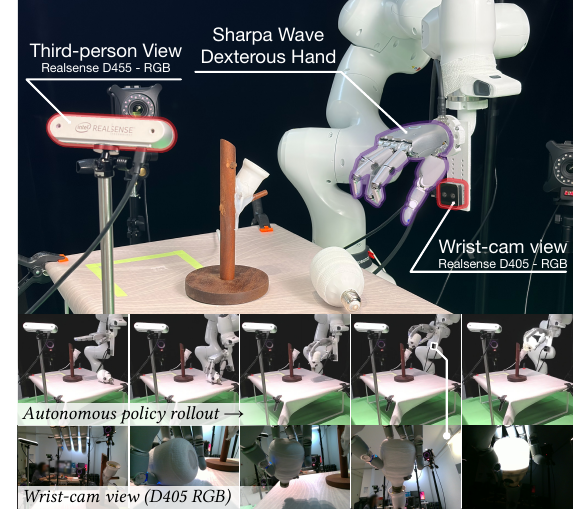}
  \vspace{-8pt}
  \caption{\textbf{Autonomous policy setup and rollout.}}
  \label{fig:auto_policy}
  \vspace{-32pt}
\end{wrapfigure}
reference motions.

\subsection{From Teleoperation to Autonomy}
\label{sec:exp:auto}

\qheading{Overview}
A key advantage of \method is its ability to collect dexterous manipulation data beyond the reach of existing teleoperation systems. While the teleoperation evaluation (\cref{sec:exp:task}) demonstrates that \method enables human-level in-hand dexterity, the collected trajectories also serve as high-quality expert demonstrations for training autonomous policies. To validate this, we train Diffusion Policies~\citep{chi2025diffusion} on three dexterous tasks: \taskBulbPolicy, \taskHammerPolicy, and \taskBrushPolicy. Each task retains the most contact-intensive stages of its teleoperation counterpart while removing the return-and-place phases, isolating the core dexterous manipulation skills.

\qheading{Protocol and Metrics}
We adopt the Conv-UNet Diffusion Policy architecture~\citep{chi2025diffusion}, conditioned on RGB observations from third-person and wrist-mounted cameras, as shown in \cref{fig:auto_policy}. For each task, we collect $50$ expert demonstrations and evaluate each policy over $15$ real-world trials. Each task is decomposed into well-defined stages following the same protocol as the teleoperation evaluation; we report stage-wise success (the number of trials surviving past each stage) and overall \SR.

\qheading{Results and Analysis}
As shown in \cref{tab:auto_policy}, all three tasks achieve non-trivial success rates from only $50$ demonstrations, confirming that \method captures sufficiently rich contact-mode coverage for behavioral cloning across diverse dexterous manipulation skills. The stage-wise breakdown reveals a distinct bottleneck per task. \taskHammerPolicy achieves the highest \SR ($73.3\%$) with perfect grasping and reorientation ($15/15$ through stage~2). The only failures occur at the nail-driving stage, where the policy must sustain repeated contact force against the foam target. \taskBulbPolicy ($46.7\%$ \SR) shows a similar pattern: grasping and reorientation succeed reliably, but the precision alignment stage ($8/12$) is the primary bottleneck. Once aligned, installation succeeds in $7/8$ trials, indicating that the screw-in skill transfers well from demonstrations. \taskBrushPolicy ($40.0\%$ \SR) presents the opposite profile: grasping is the dominant failure point ($7/15$), due to the thin and irregular brush handle requiring precise finger placement that is difficult to resolve from RGB alone. Trials that survive the grasp nearly all complete the subsequent rotation and sweep ($6/7$).

Notably, no baseline teleoperation system evaluated in \cref{tab:task_results} can reliably complete any of these three tasks, making it infeasible to collect comparable demonstration data with existing methods. These results validate \method as both a teleoperation interface and a scalable data collection pipeline for autonomous dexterous manipulation.

\subsection{Ablation Studies}
\label{sec:exp:ablation}
\qheading{Consecutive Subgoal Tracking vs.\ Dense Tracking}
We compare our consecutive subgoal tracking formulation against standard dense frame-wise tracking in simulation on held-out reference motions for three objects (\cref{tab:ablation_subgoal}). We evaluate in two modes: dense, where the reference advances every control step, and sparse, where the target advances only after the current subgoal is reached. Even under dense evaluation, which favors the dense tracking baseline, sparse subgoal tracking achieves significantly longer episode lengths. Dense frame-wise tracking forces the policy to replicate reference trajectories step by step, leaving insufficient tolerance to discover physically feasible contact strategies and causing early termination. In contrast, sparse subgoal tracking only requires stable goal completion, allowing the policy to find feasible contact sequences through simulation rollout. This advantage is amplified in sparse evaluation, where dense tracking policies stall after only a few subgoals while sparse subgoal tracking reaches orders of magnitude more.

\qheading{Random Action Masking}
We ablate random action masking on three real-world teleoperation tasks (\cref{tab:ablation_am}). Removing action masking causes substantial degradation across all evaluated tasks. Random action masking serves as a strong action-space regularization that prevents the policy from overfitting to simulation dynamics, which inevitably differ from real-world dynamics. Without this regularization, the policy exploits simulation-specific dynamics patterns that do not transfer, confirming that random action masking is critical for zero-shot sim-to-real deployment.

\begin{table*}[!t]
    \vspace{-8pt}
    \centering
    \captionsetup{font=small}
    \begin{minipage}[t]{0.52\linewidth}
        \centering
        \caption{\textbf{Sparse subgoal vs.\ dense tracking} (sim). EpLen: episode length ($\uparrow$). Goals: consecutive subgoals reach ($\uparrow$).}
        \label{tab:ablation_subgoal}
        \vspace{-6pt}
        \footnotesize
        \setlength{\tabcolsep}{3pt}
        \begin{tabular}{lcccc}
            \toprule
            & \multicolumn{2}{c}{Dense eval: EpLen} & \multicolumn{2}{c}{Sparse eval: Goals} \\
            \cmidrule(lr){2-3} \cmidrule(lr){4-5}
            Object & Ours & Dense & Ours & Dense \\
            \midrule
            Cuboid      & \textbf{378.6} & 115.8 & \textbf{32.6}  & 2.6 \\
            Hammer      & \textbf{376.9} & 131.2 & \textbf{186.6} & 2.7 \\
            Screwdriver & \textbf{373.2} &  88.7 & \textbf{178.5} & 2.7 \\
            \bottomrule
        \end{tabular}
    \end{minipage}%
    \hfill
    \begin{minipage}[t]{0.44\linewidth}
        \centering
        \caption{\textbf{Random action masking} (real). Each cell reports \SR / \TP (\%).}
        \label{tab:ablation_am}
        \footnotesize
        \setlength{\tabcolsep}{3pt}
        \begin{tabular}{lcc}
            \toprule
            Task & w/ AM & w/o AM \\
            \midrule
            \taskHammer      & \textbf{66.7} / \textbf{86.7} & 33.3 / 57.1 \\
            \taskScrewdriver & \textbf{73.3} / \textbf{86.7} &  0.0 / 36.0 \\
            \taskCuboid      & \textbf{80.0} / \textbf{86.7} & 26.7 / 51.1 \\
            \bottomrule
        \end{tabular}
    \end{minipage}
    \vspace{-8pt}
\end{table*}

\section{Conclusion}
\label{sec:conclusion}
We introduce \method, a hand-object co-tracking controller that takes a concrete step toward human-level dexterous teleoperation. Built on our proposed \emph{consecutive subgoal tracking} with \emph{hybrid reward design}, and \emph{random action masking}, \method learns diverse in-hand skills in single-stage \ac{rl} with no task-specific reward, and transfers to real robots. Across seven long-horizon and challenging dexterous tasks, \method achieves a strong success rate where baselines uniformly fail. Furthermore, the collected demonstrations successfully train policies to autonomously execute long-horizon dexterous tasks, establishing \textbf{a scalable foundation for collecting rich in-hand dexterous manipulation data and a viable path toward human-level robotic dexterity}.

\section{Limitations}
\method currently learns an object-specific controller. Adapting to a new object requires collecting human hand-object interaction data and training a dedicated policy. Scaling to a unified, object-conditioned controller that generalizes across object categories without per-object data collection and training is a promising direction for future work.

Our real-world deployment relies on a motion-capture system for real-time hand and object pose estimation. Replacing this with a markerless, vision-based tracking system would significantly lower the barrier to deployment and broaden the practical applicability of the framework.

\clearpage
\bibliography{reference}

\newpage
\appendix
\onecolumn
\UseRawInputEncoding
\clearpage

\raggedbottom  %

\begin{center}
    \large\textbf{Supplementary Materials of} \\
    \large\textbf{Towards Human-level Dexterous Teleoperation}
\end{center}

This appendix complements the main paper with extended results and full implementation details.
\cref{appx:results} provides extended results and analysis, including any-to-any reposition tests, additional ablation studies, stage-wise teleoperation analysis, and failure modes.
\cref{appx:setup} details the task definitions, hardware setup, and teleoperation interface.
\cref{appx:method} details the complete implementation of \method.
\cref{appx:baselines} details each baseline implementation, and \cref{appx:auto} details the autonomous policy architecture and training.

\section{Extended Results and Analysis}
\label{appx:results}

\subsection{Any-to-Any Reposition Evaluation}
\label{appx:results:any2any}
\label{appx:cotrack:any2any}  %

\qheading{Overview}
This evaluation directly tests whether our co-tracking controller, deployed zero-shot to the real robot, can handle arbitrary in-hand hand-object tracking goal transitions. During teleoperation the operator's goal can jump to any feasible hand-object configuration at any time; the any-to-any reposition test isolates this capability by streaming a sequence of randomly sampled targets and measuring how many the controller reaches consecutively before failure.

\qheading{Protocol and Metrics}
We evaluate the co-tracking controller on \texttt{Cylinder} and \texttt{Cuboid} on LeapHand. We choose LeapHand for this stress test. Despite LeapHand's more limited dexterity ($4$ fingers, $16$~DoFs vs.\ $5$ fingers, $22$~DoFs), the controller trained by our framework still demonstrates non-trivial in-hand repositioning capability and robustness, as shown below.

For each object, we construct a hand-object targets pool consisting of all frames from our HOI reference set (\cref{appx:method:hoi}), which covers the full feasible workspace of in-hand configurations. We run $15$ trials per object. Each trial begins by sampling an initial grasp pose from the pool; the tester places the object into the hand accordingly and starts the controller. The controller is then queried with a stream of targets, each freshly sampled from the same pool as soon as the previous one is reached. A target is counted as reached when the object position error falls below $2$\,cm and the object rotation error falls below $20^{\circ}$ simultaneously. A trial terminates when (i)~the object slips out of the hand, or (ii)~the active target has not been reached for $20$\,s. We report two metrics: consecutive successes, the number of targets reached in sequence before failure (averaged over trials), and per-target success rate, the fraction of all attempted targets that are successfully reached.

\qheading{Results and Analysis}
As shown in \cref{tab:any2any}, the controller sustains long sequences of arbitrary goal transitions on both objects, confirming that it generalizes well beyond the training reference trajectories. On \texttt{Cylinder} the controller reaches $41.1$ consecutive targets on average with a per-target success rate of $97.6\%$, substantially outperforming \texttt{Cuboid} ($12.1$ consecutive successes, $92.3\%$). We attribute this gap to object geometry: the cylinder's continuous surface allows fingers to slide and roll the object fluidly during transitions without encountering abrupt geometric changes, and its rotational symmetry reduces the effective distance between arbitrary target poses. The cuboid's edges and corners, by contrast, can obstruct finger motion during rapid regrasps --- fingers must navigate around these features to reach certain target orientations, increasing the chance of contact transition jams and failed repositioning attempts.

\begin{table}[!h]
\vspace{-12pt}
    \centering
    \captionsetup{font=small}
    \caption{\textbf{Any-to-any results on LeapHand.}
    Each object is evaluated over $15$ trials.}
    \label{tab:any2any}
    \vspace{4pt}
    \setlength{\tabcolsep}{8pt}
    \begin{tabular}{lcc}
    \toprule
    \textbf{Object} & Consecutive Successes ($\uparrow$) & Per-target SR ($\uparrow$) \\
    \midrule
    \texttt{Cylinder} & $41.1$ & $97.6\%$ \\
    \texttt{Cuboid}   & $12.1$ & $92.3\%$ \\
    \bottomrule
    \end{tabular}
\end{table}

\subsection{Additional Ablation Studies}
\label{appx:results:ablations}

\begin{figure}[!h]
    \centering
    \begin{minipage}[t]{0.48\linewidth}
        \centering
        \includegraphics[width=\linewidth]{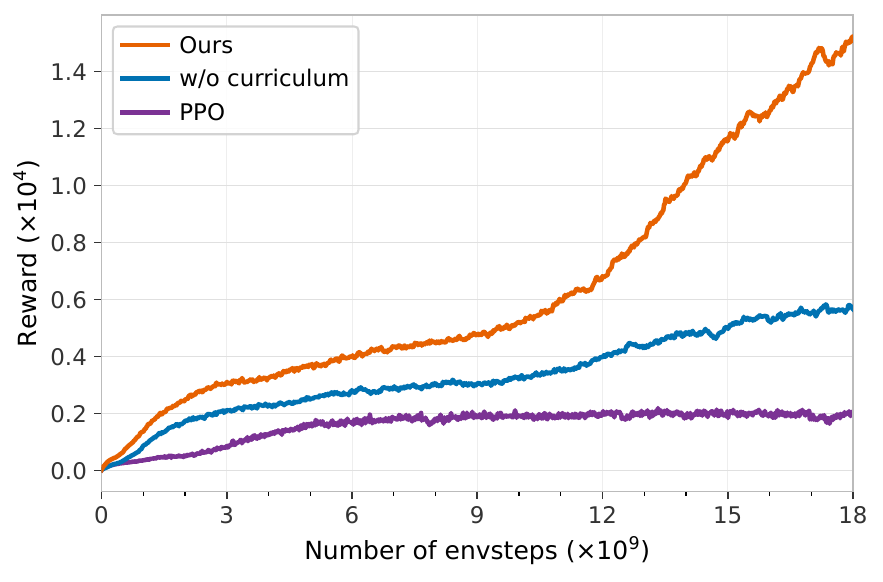}\\[2pt]
        {\small (a) Reward.}
    \end{minipage}\hfill
    \begin{minipage}[t]{0.48\linewidth}
        \centering
        \includegraphics[width=\linewidth]{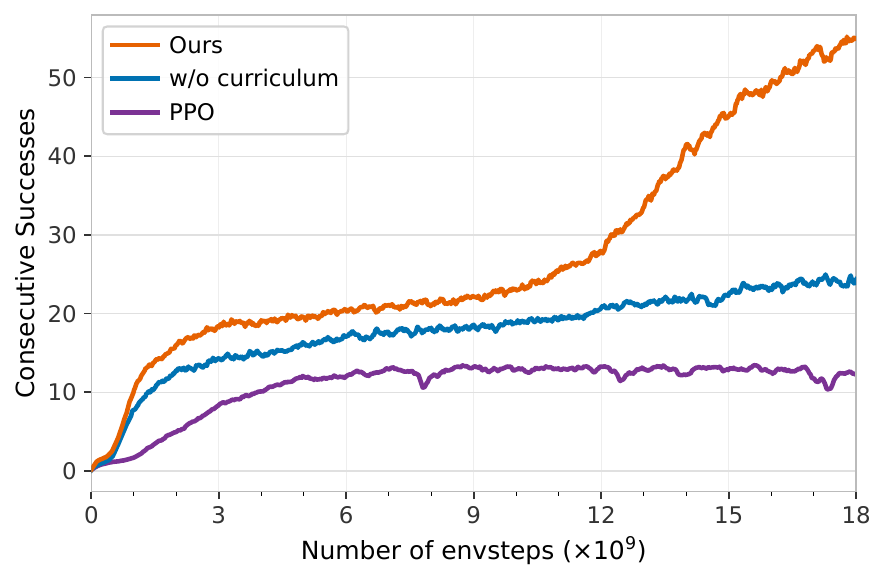}\\[2pt]
        {\small (b) Consecutive successes.}
    \end{minipage}
    \caption{\textbf{Training-recipe ablation curves on \texttt{Hammer}.} (a)~Reward and (b)~consecutive subgoals reached vs.\ environment steps for the full method, without curriculum, and with PPO replacing SAPG.}
    \label{fig:ablation_curves}
\end{figure}

We ablate two training recipe choices on \texttt{Hammer}: the curriculum schedule (\cref{appx:method:curriculum}) and the policy optimizer (SAPG vs.\ PPO). All other training settings and reference motions are shared. We report training curves of reward and consecutive subgoals reached in \cref{fig:ablation_curves}.

\qheading{Curriculum Schedule}
As shown in \cref{fig:ablation_curves}, disabling the curriculum (training at full deployment difficulty throughout) produces faster initial progress but plateaus at a lower final performance. The curriculum completes its annealing within the first $2 \times 10^{9}$ environment steps, yet the advantage it provides continues to grow well beyond that point. We attribute this to the quality of the early-training foundation: the curriculum first exposes the policy to small subgoal steps, permissive tolerances, and reduced gravity, allowing it to discover a diverse repertoire of stable contact primitives before difficulty ramps up. Without this scaffolding, the policy must simultaneously learn basic contact strategies and cope with full-difficulty dynamics, converging to a narrower set of behaviors that limits its ability to compose longer manipulation sequences later in training.

\qheading{PPO vs.\ SAPG}
As shown in \cref{fig:ablation_curves}, replacing SAPG~\citep{singla2024sapg} with vanilla PPO causes a substantial drop in both reward and consecutive successes. SAPG maintains multiple independently exploring policy blocks whose gradients are aggregated, promoting diverse strategy discovery within a single training run. This exploration diversity is critical for co-tracking, where the policy must master qualitatively different contact modes --- translation, continuous rotation, finger gaiting, and regrasping --- within a single network. PPO's unimodal gradient updates tend to commit early to a limited strategy set, under-covering the full spectrum of in-hand manipulation modalities in our reference motions.

\subsection{Stage-Wise Teleoperation Success Analysis}
\label{appx:results:stagewise}

\begin{figure}[!h]
    \centering
    \includegraphics[width=\linewidth]{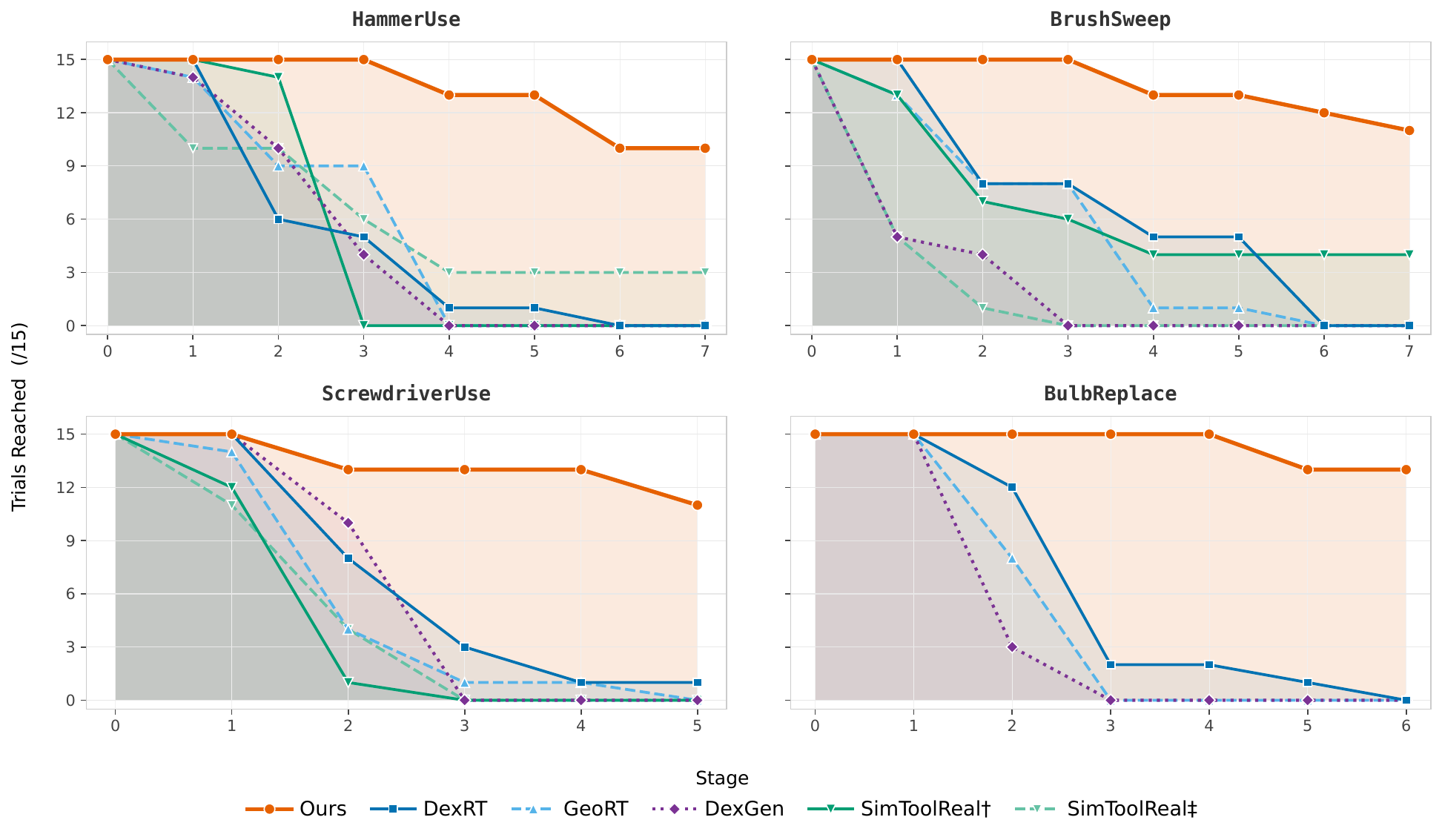}
    \caption{\textbf{Stage-wise success on four long-horizon tool-use tasks.} For each task, the horizontal axis indexes the task stage and the vertical axis is the number of trials (out of $15$) that reach that stage. Each curve corresponds to one method (\method and baselines from \cref{tab:task_results} in the main paper). \textit{SimToolReal} is not a teleoperation method but is included as a strong baseline for learned tool manipulation.}
    \label{fig:stagewise_real}
\end{figure}

For each of the four long-horizon tool-use tasks, we plot the number of trials (out of~$15$) that survive past each task stage for \method and all baselines (\cref{fig:stagewise_real}). The main paper highlights key stage-wise numbers for \method; here we provide the complete per-stage breakdown across all methods and analyze where each baseline breaks down. We additionally include \textit{SimToolReal}~\citep{kedia2026simtoolreal}, an object-centric sim-to-real tool-use policy that is not a teleoperation method but provides a strong reference for learned dexterous tool manipulation.

Across all four tasks, a consistent pattern emerges. \method retains $13$--$15$ out of $15$ trials through the demanding mid-task stages and only drops modestly at the final placement stages, whereas every baseline suffers a sharp collapse at or shortly after the first stage that requires in-hand reorientation or functional grasp transition.

In \taskHammer, all methods pick up the hammer (stage~$1$), but baselines diverge sharply at stage~$2$ (rotate face-down): \textit{DexRT} drops from $15$ to $6$ and \textit{DexGen} from $14$ to $10$. By stage~$4$ (rotate claw-down), no teleoperation baseline retains any trial. \textit{SimToolReal}$^\dagger$ (category-specific) leverages task-specific training to reach $14/15$ at stage~$2$ but collapses entirely at stage~$3$ (drive nails), unable to sustain the repeated contact force required for hammering. \textit{SimToolReal}$^\ddagger$ (all-category) maintains $3/15$ through completion at the cost of weaker early-stage performance. \method retains all $15$ trials through stage~$3$ and $10$ through final placement.

In \taskBrush, a similar bottleneck emerges at grasp-transition stages. \textit{DexGen} collapses earliest, losing most trials at pick-up ($5/15$) due to compounding generative-model errors. \textit{DexRT} and \textit{GeoRT} survive the initial sweep (stage~$3$, $8/15$ each) but fail at stage~$4$ (rotate bristles-right), which demands controlled in-hand reorientation while maintaining grasp. \textit{SimToolReal}$^\dagger$ is the strongest baseline on this task with $4/15$ completions, while \method reaches $11/15$.

In \taskScrewdriver, \textit{DexGen} maintains $10/15$ through stage~$2$ (rotate to align) but drops to $0$ at stage~$3$ (tighten the screw), as its action prior cannot sustain continuous axial rotation. Both \textit{SimToolReal} variants also fail entirely at stage~$3$. \textit{DexRT} retains $1/15$ to completion, the only teleoperation baseline trial to finish this task. \method maintains $13/15$ through tightening and finishes with $11/15$.

In \taskBulb, evaluated without \textit{SimToolReal} as the task falls outside its object-centric formulation, all methods pick up the bulb ($15/15$), but the critical drop occurs at stage~$3$ (screw in), where \textit{DexRT} falls from $12$ to $2$ and both \textit{GeoRT} and \textit{DexGen} reach $0$. These stages require precise bidirectional rotation about the bulb axis under sustained contact, which kinematic retargeting cannot achieve. \method completes both rotational stages without trial loss ($15/15$ through stage~$4$) and finishes with $13/15$.

Taken together, these results show that all baselines fail not at grasping or gross positioning, but at contact-rich in-hand manipulation: reorientation, finger gaiting, and sustained tool application. Kinematic methods (\textit{DexRT}, \textit{GeoRT}) lack a dynamics prior and collapse at the first contact-intensive stage. \textit{DexGen} suffers compounding trajectory drift that causes abrupt failure at contact transitions. \textit{SimToolReal} achieves partial success on trained tool categories but cannot generalize across the diverse contact modes within a single long-horizon task. \method sustains high trial survival precisely at these stages, confirming that the co-tracking controller provides the in-hand dexterity needed for long-horizon tool use.

\subsection{\method Failure Analysis}
\label{appx:results:failure}
\label{appx:cotrack:failure}  %

\begin{figure}[!t]
    \centering
    \includegraphics[width=\linewidth]{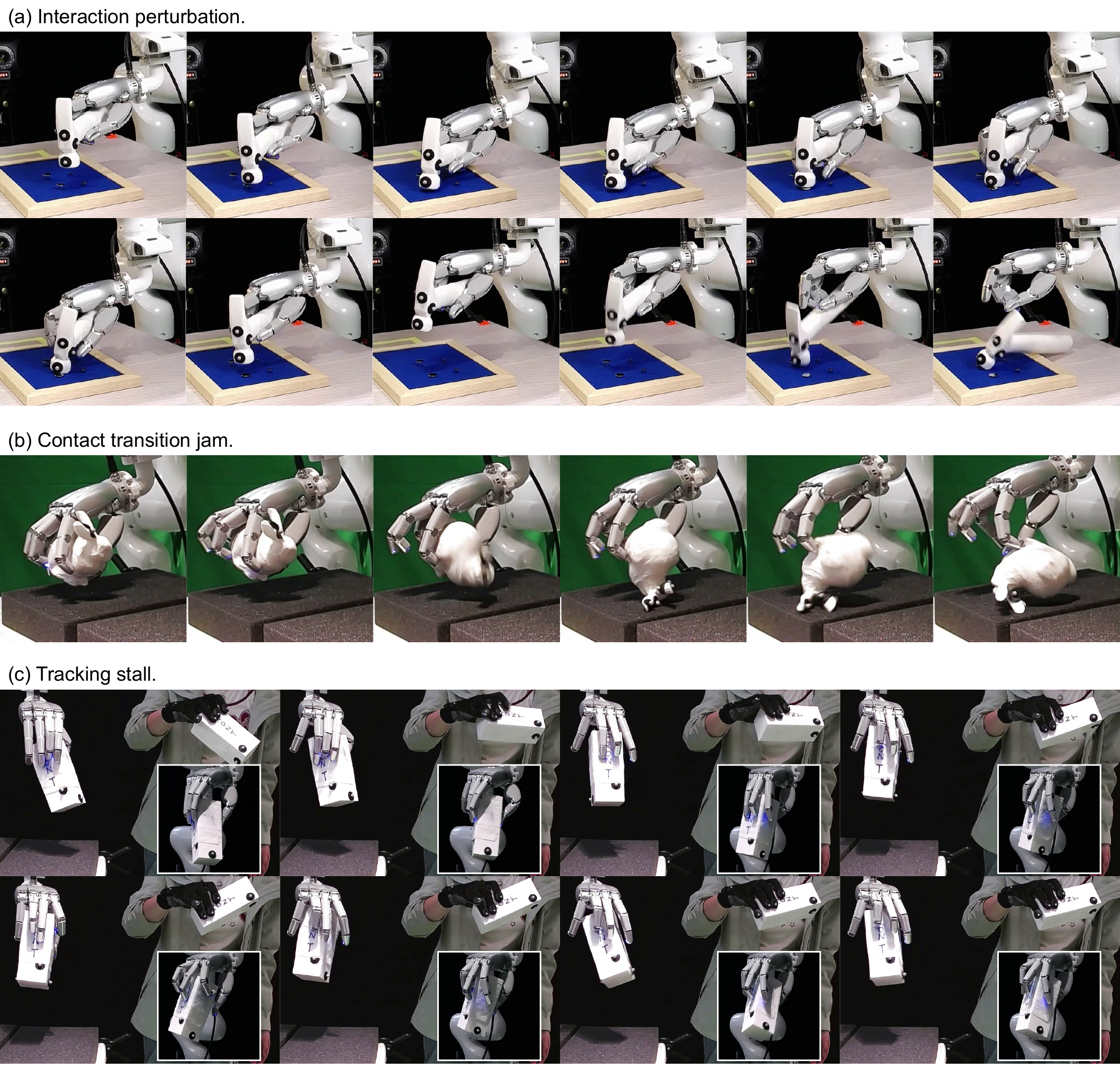}
    \vspace{-18pt}
    \caption{\textbf{Real-world failure cases.}
    Each panel shows a snapshot at the point of failure together with the representative failure mode:
    (a)~interaction perturbation,
    (b)~contact transition jam,
    (c)~tracking stall.}
    \label{fig:failure_real}
    \vspace{-12pt}
\end{figure}

We identify three dominant failure modes of our
teleoperation controller, illustrated in \cref{fig:failure_real}.

\begin{itemize}[leftmargin=2em,noitemsep,nolistsep,topsep=2pt,partopsep=0pt]

\item \textbf{Interaction perturbation (\cref{fig:failure_real}a).} Forceful tool--environment contact, such as striking a nail during \taskHammer, generates impulsive reaction forces that shift the object's in-hand pose by a large, instantaneous amount. The controller, trained exclusively on free-space hand--object interaction, has never encountered such impact dynamics and cannot recover from the resulting out-of-distribution grasp configuration. This mode does not appear in reorientation-only tasks.

\item \textbf{Contact transition jam (\cref{fig:failure_real}b).} During finger gaiting or in-hand reorientation, a finger that should lift away occasionally remains wedged against the object due to actuator compliance or geometric interlocking. As the remaining fingers continue to move, the jam generates unbalanced forces that eject the object from the hand. This failure is most frequent on irregular objects (\taskBunny) where concavities increase the chance of interlocking.

\item \textbf{Tracking stall (\cref{fig:failure_real}c).} The controller attempts a regrasp but fails to establish contacts that can move the object toward the target pose. Unlike the contact transition jam, the object is not lost---the hand simply cannot make progress. We attribute this to the absence of tactile observations (\cref{tab:obs_dims}): the policy cannot distinguish between a finger pressing against the object and one sliding past it, and thus cannot adapt when a regrasp attempt fails.

\end{itemize}

The three modes point to distinct limitation of our current system. Interaction perturbation reflects a \emph{training distribution} gap, as tool--environment impact dynamics are absent from the simulation training. Contact transition jams reveal an \emph{actuation compliance} mismatch between rigid-body simulation and real direct-drive fingers with passive compliance. Tracking stalls expose the lack of \emph{tactile observation}, without which the policy cannot close the loop on contact state during regrasping. Addressing these limits through interaction-aware training, compliant-contact simulation, and tactile-rich observation spaces is a promising direction for future work.

\section{Experimental Setup Details}
\label{appx:setup}

\subsection{Teleoperation Task Definitions}
\label{appx:setup:tasks}

We provide detailed descriptions of the seven real-world tasks introduced in the main paper (\cref{fig:task_description}), together with their stage decompositions. Each task is decomposed into $N$ well-defined stages that the trial must reach in order. The task progress (\TP) reported in the main paper is the average $k/N$ of the furthest stage $k$ reached across trials, and the success rate (\SR) is the fraction of trials that reach all $N$ stages. For all tasks, stage $0/N$ denotes failure to pick up the object. We wrap the tested objects with medical bandage tape to increase surface friction.

\textbf{\taskCylinder\,/\,\taskCuboid\,/\,\taskBunny\ ($3$ stages each).}
The three reorientation tasks share an identical stage decomposition and differ only in object geometry, testing continuous in-hand pose control on rotationally symmetric (\taskCylinder), corner-rich (\taskCuboid), and irregular freeform (\taskBunny) geometries.
\emph{Stages:}
\begin{itemize}[leftmargin=2em,noitemsep,nolistsep,topsep=0pt,partopsep=0pt]
  \item $1/3$: picked up the object.
  \item $2/3$: reoriented to the target pose in-hand.
  \item $3/3$: placed back on the table.
\end{itemize}

\textbf{\taskHammer\ ($7$ stages).}
A long-horizon task that exercises bidirectional functional grasp transitions (face-down vs.\ claw-down) and repeated striking and pulling.
\emph{Stages:}
\begin{itemize}[leftmargin=2em,noitemsep,nolistsep,topsep=0pt,partopsep=0pt]
  \item $1/7$: picked up the hammer.
  \item $2/7$: rotated face-down for hammering.
  \item $3/7$: drove the two nails into the board.
  \item $4/7$: rotated claw-down.
  \item $5/7$: pulled out the two nails with the claw.
  \item $6/7$: rotated parallel to the table.
  \item $7/7$: placed back on the table.
\end{itemize}

\textbf{\taskBrush\ ($7$ stages).}
A long-horizon task requiring transitions between two functional brush orientations while sweeping across an extended workspace.
\emph{Stages:}
\begin{itemize}[leftmargin=2em,noitemsep,nolistsep,topsep=0pt,partopsep=0pt]
  \item $1/7$: picked up the brush.
  \item $2/7$: rotated for forward sweeping.
  \item $3/7$: swept the debris forward.
  \item $4/7$: rotated so the bristles face rightward.
  \item $5/7$: swept the debris rightward into the target area.
  \item $6/7$: rotated parallel to the table.
  \item $7/7$: placed back on the table.
\end{itemize}

\textbf{\taskScrewdriver\ ($5$ stages).}
A precision task that tests axial alignment with the screw and sustained in-hand rotation about the tool axis.
\emph{Stages:}
\begin{itemize}[leftmargin=2em,noitemsep,nolistsep,topsep=0pt,partopsep=0pt]
  \item $1/5$: picked up the screwdriver.
  \item $2/5$: rotate the screwdriver tip downward..
  \item $3/5$: tightened the screw until fully seated.
  \item $4/5$: rotated the screwdriver for placement.
  \item $5/5$: placed back on the table.
\end{itemize}

\textbf{\taskBulb\ ($6$ stages).}
A precision insertion task that combines functional in-hand reorientation with bidirectional rotation about the bulb axis.
\emph{Stages:}
\begin{itemize}[leftmargin=2em,noitemsep,nolistsep,topsep=0pt,partopsep=0pt]
  \item $1/6$: picked up the bulb.
  \item $2/6$: rotated to align with the socket.
  \item $3/6$: screwed the bulb in until it lit up.
  \item $4/6$: unscrewed it until fully removed.
  \item $5/6$: rotated for placement.
  \item $6/6$: placed back on the table.
\end{itemize}

\subsection{Autonomous Policy Task Definitions}
\label{appx:setup:auto_tasks}

The autonomous Diffusion Policies are evaluated on simplified
subsets of three teleoperation tasks, each retaining the most
dexterous stages while removing the return-and-place phases.
Stage decompositions follow the same protocol as the teleoperation
tasks (\cref{appx:setup:tasks}); a trial is terminated when an
unrecoverable failure occurs.

\textbf{\taskBulbPolicy\ ($4$ stages).}
A subset of \taskBulb that covers bulb installation only
(no unscrewing or return).
\emph{Stages:}
\begin{itemize}[leftmargin=2em,noitemsep,nolistsep,topsep=0pt,partopsep=0pt]
  \item $1/4$: picked up the bulb.
  \item $2/4$: reoriented to the installation pose.
  \item $3/4$: aligned with the socket.
  \item $4/4$: screwed in until the bulb lit up.
\end{itemize}

\textbf{\taskHammerPolicy\ ($3$ stages).}
A simplified variant of \taskHammer with larger nails and a foam
target, retaining the core in-hand rotation to a functional
hammering grasp.
\emph{Stages:}
\begin{itemize}[leftmargin=2em,noitemsep,nolistsep,topsep=0pt,partopsep=0pt]
  \item $1/3$: picked up the hammer.
  \item $2/3$: rotated face-down for hammering.
  \item $3/3$: drove the nails into foam.
\end{itemize}

\textbf{\taskBrushPolicy\ ($3$ stages).}
A subset of \taskBrush that covers a single forward sweep only
(no bristle-reorientation or return sweep).
\emph{Stages:}
\begin{itemize}[leftmargin=2em,noitemsep,nolistsep,topsep=0pt,partopsep=0pt]
  \item $1/3$: picked up the brush.
  \item $2/3$: rotated for forward sweeping.
  \item $3/3$: swept forward across the workspace.
\end{itemize}

\subsection{Hardware and Teleoperation System}
\label{appx:setup:hardware}

\subsubsection{Motion Capture System}
\label{appx:setup:mocap}
\label{appx:hardware:capture}  %

\begin{figure}[!h]
    \centering
    \begin{minipage}[t]{0.48\linewidth}
        \centering
        \includegraphics[width=\linewidth]{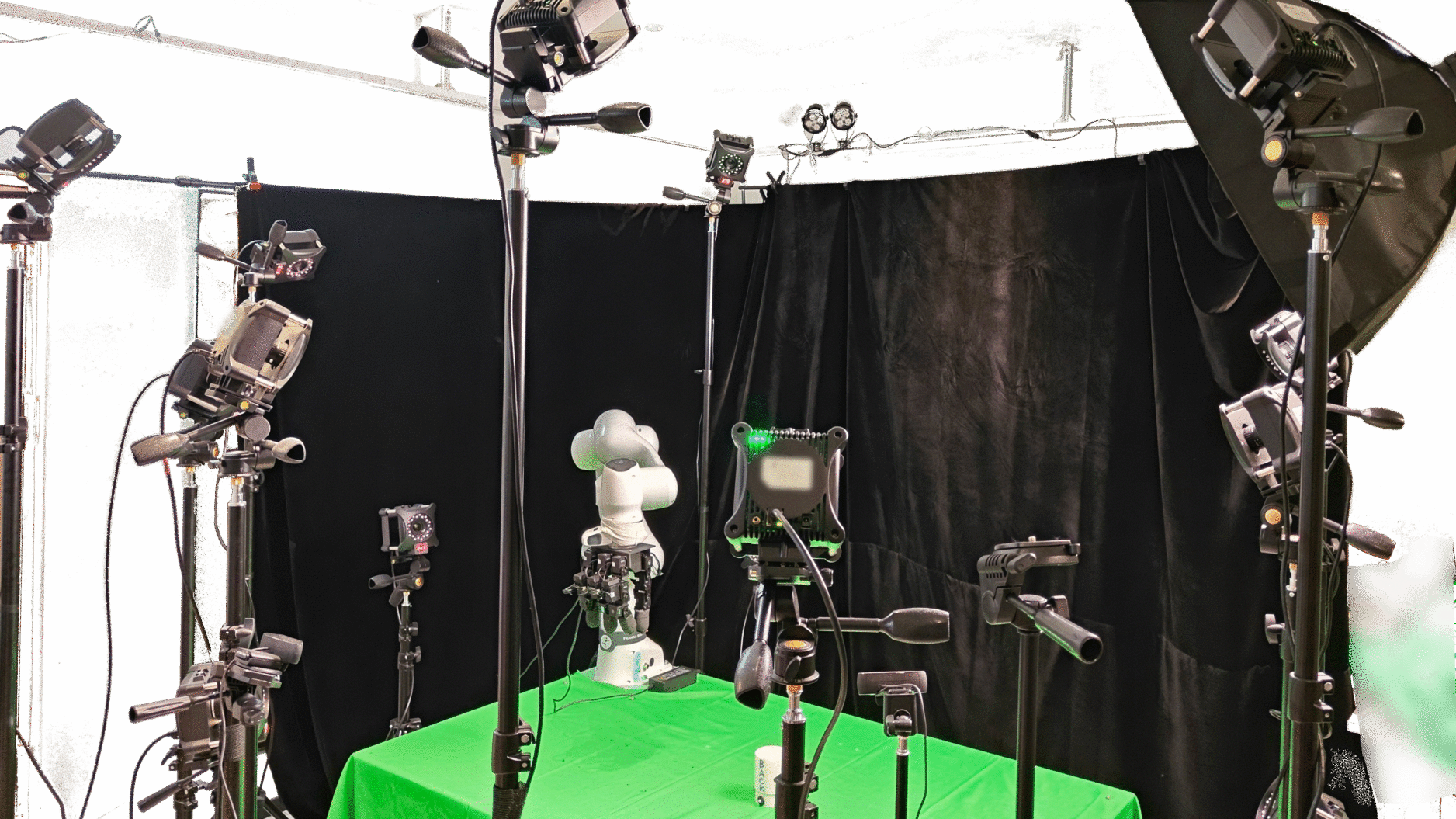}\\[2pt]
        {\small (a) Motion capture setup.}
    \end{minipage}\hfill
    \begin{minipage}[t]{0.48\linewidth}
        \centering
        \includegraphics[width=\linewidth]{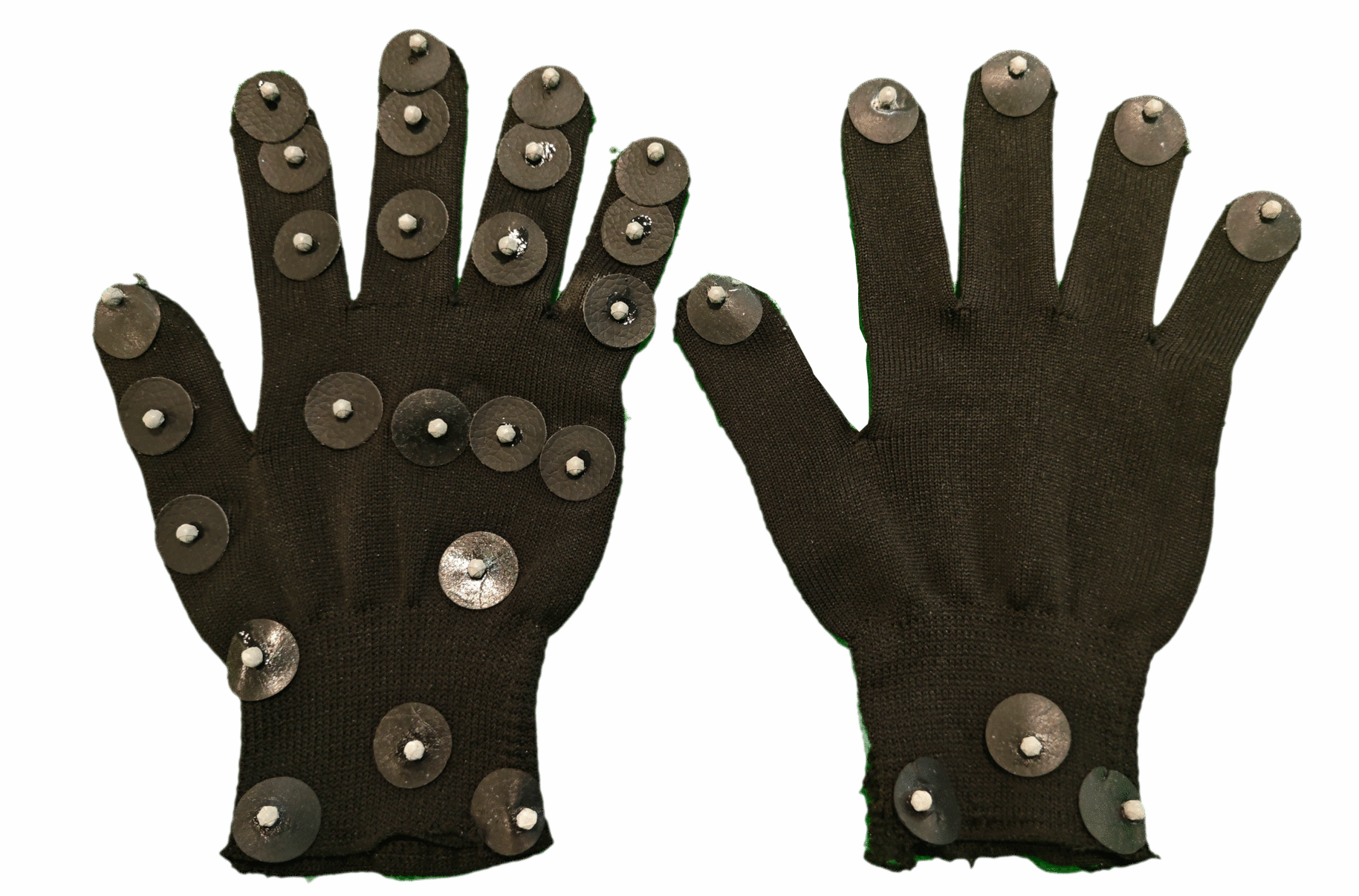}\\[2pt]
        {\small (b) Marker-instrumented gloves.}
    \end{minipage}
    \caption{\textbf{Hand--object motion capture setup.} (a)~The dedicated capture volume equipped with tripod-mounted NOKOV infrared cameras. (b)~Two glove configurations: the left glove, used for offline reference-motion collection, has dense markers on the wrist, palm, and all finger joints; the right glove, used for real-time teleoperation, has markers only on the wrist and fingertips.}
    \label{fig:mocap_setup}
\end{figure}

We use a NOKOV optical motion capture system for both offline reference-motion collection and real-time teleoperation, with a different glove configuration for each setting. For offline collection (\cref{fig:mocap_setup}a), the operator stands in a dedicated capture volume equipped with tripod-mounted infrared cameras and wears a full-coverage glove instrumented with retro-reflective markers on the wrist, palm, and all finger joints (\cref{fig:mocap_setup}b, left). This dense marker layout enables the system to output, per frame, (i)~$24$ hand joint $3$-D positions, (ii)~$21$ joint pose matrices, (iii)~the wrist $\mathrm{SE}(3)$ pose, and (iv)~the object's $6$-D pose. For real-time teleoperation, the operator wears a compact glove with markers placed only on the wrist and fingertips (\cref{fig:mocap_setup}b, right) at the deployment workstation. This lightweight configuration provides the wrist pose and fingertip positions needed by the control loop. The MoCap stream is fed directly into the teleoperation loop at $30$\,Hz. In both settings, each rigid object is tagged with a marker rigid body that yields its $6$-D pose.

\subsubsection{Teleoperation Interface}
\label{appx:setup:teleop}
\label{appx:hardware:teleop}  %

At runtime, the NOKOV system (\cref{appx:setup:mocap}) streams the operator's wrist pose and fingertip positions together with the manipulated object's $6$-D pose at $30$\,Hz. Each frame triggers two parallel control paths that together close the teleoperation loop at the same rate:
\begin{itemize}[leftmargin=*,noitemsep,nolistsep,topsep=0pt,partopsep=0pt]
  \item \textbf{Arm.} The operator's wrist pose (world frame) is mapped to a $7$-DoF Franka FR3 joint target via inverse kinematics.
  \item \textbf{Hand.} The operator's fingertip positions and the object's $6$-D pose are converted to the wrist frame and assembled into the co-tracking goal $g_t$. The learned policy then maps the current hand--object state and $g_t$ to joint position targets sent to the dexterous-hand SDK via position control.
\end{itemize}

The co-tracking policy is trained on in-hand HOI references (\cref{appx:method:hoi}) and does not cover the pre-grasp approach. We therefore split each trial into two phases. In the \emph{reaching and grasping phase}, the hand is driven by kinematic vector retargeting~\citep{qin2023anyteleop,handa2020dexpilot}, which maps the operator's fingertip positions to robot joints via vector alignment and handles the pre-grasp approach and initial contact acquisition. Once the hand reaches an approximate grasp pose near the object, the operator switches to the \emph{in-hand phase}, where the co-tracking policy drives all subsequent in-hand reorientation, regrasping, and finger gaiting.

\section{\method Implementation Details}
\label{appx:method}

This section provides the full implementation details of \method. We follow the same notation as \cref{sec:method} unless stated otherwise.

\subsection{Observation \& Action Space}
\label{appx:method:arch}

\qheading{Observation Space}
The observation $\mbold{o}_t$ concatenates the elements listed in \cref{tab:obs_dims} (SharpaWave: $N_f=5$, $n_{\mathrm{dof}}=22$; LeapHand: $N_f=4$, $n_{\mathrm{dof}}=16$). Two design choices are worth highlighting: (i)~since every other entry is expressed in $\mathcal{F}_{\mathrm{wrist}}$, the gravity direction $\hat{\mbold{g}}_{\mathrm{wrist}}$ is the \emph{only} signal that anchors the policy to the world frame and tells it the absolute orientation of the wrist; and (ii)~we deliberately exclude joint velocities and contact forces, since both are noisy or unavailable on hardware and the policy can implicitly recover them from the kinematic state, previous action, and gravity cue.

\begin{table}[!h]
\vspace{-12pt}
\centering
\captionsetup{font=small}
\caption{\textbf{Observation $\mbold{o}_t$: per-element dimensions.}
$\hat{\cdot}$~denotes a target quantity;\; $\Delta$~denotes target $-$ current.}
\label{tab:obs_dims}
\small
\setlength{\tabcolsep}{6pt}
\begin{tabular}{llccc}
\toprule
Element & Description & Formula & SharpaWave & LeapHand \\
\midrule
$\mbold{q}_t$                              & joint positions                       & $n_{\mathrm{dof}}$  & $22$  & $16$  \\
$\cos\mbold{q}_t$                          & cosine of joint positions             & $n_{\mathrm{dof}}$  & $22$  & $16$  \\
$\sin\mbold{q}_t$                          & sine of joint positions               & $n_{\mathrm{dof}}$  & $22$  & $16$  \\
\midrule
$\mbold{x}^o_t$                            & object position in $\mathcal{F}_{\mathrm{wrist}}$         & $3$                 & $3$   & $3$   \\
$\mbold{q}^o_t$                            & object quaternion in $\mathcal{F}_{\mathrm{wrist}}$       & $4$                 & $4$   & $4$   \\
\midrule
$\hat{\mbold{g}}_{\mathrm{wrist}}$         & gravity direction in $\mathcal{F}_{\mathrm{wrist}}$       & $3$                 & $3$   & $3$   \\
\midrule
$\hat{\mbold{p}}^{\mathrm{tip}}_t$         & target fingertip positions             & $3 N_f$             & $15$  & $12$  \\
$\Delta\hat{\mbold{p}}^{\mathrm{tip}}_t$   & fingertip target $-$ current           & $3 N_f$             & $15$  & $12$  \\
$\hat{\mbold{x}}^o_t$                      & target object position                 & $3$                 & $3$   & $3$   \\
$\Delta\hat{\mbold{x}}^o_t$                & object-position target $-$ current     & $3$                 & $3$   & $3$   \\
$\hat{\mbold{q}}^o_t$                      & target object quaternion               & $4$                 & $4$   & $4$   \\
$\Delta\hat{\mbold{q}}^o_t$                & object-quaternion target $-$ current   & $4$                 & $4$   & $4$   \\
$\tilde{\mbold{a}}_{t-1}$                  & previous low-level command             & $n_{\mathrm{dof}}$  & $22$  & $16$  \\
\midrule
\textbf{Total} $|\mbold{o}_t|$             &                                       &                     & $\mathbf{142}$ & $\mathbf{112}$ \\
\bottomrule
\end{tabular}
\end{table}

\qheading{Action Space}
Extending the residual joint-target parameterization of HORA~\citep{qi2023inhand} with a soft deadzone, the policy output $\mbold{a}_t \in [-1, 1]^{n_{\mathrm{dof}}}$ is converted into a joint command by
\[
\mbold{a}^{\mathrm{cmd}}_t = \mathrm{clip}_{\mathrm{joint}}\!\Bigl(\mbold{a}^{\mathrm{cmd}}_{t-1} + \alpha_a \cdot \mathrm{deadzone}_{\tau}(\mbold{a}_t)\Bigr),
\quad
\mathrm{deadzone}_\tau(a) = \mathrm{sign}(a)\,\max\!\bigl(|a|-\tau,\,0\bigr),
\]
with action scale $\alpha_a = 0.1$, deadzone threshold $\tau = 0.1$. Under this residual formulation, exactly outputting zero is hard for a Gaussian policy; the deadzone gives an explicit ``hold-still'' region in action space (any $\mbold{a}_t$ within $\pm\tau$ produces zero delta), letting the policy actively choose to keep the current command rather than having to emit exactly zero.

\subsection{Complete Reward Design}
\label{appx:method:reward}

We give the complete form of the reward used to train the co-tracking controller. The reward couples a sparse \emph{consecutive subgoal-reaching} term, a dense \emph{tracking} term, and a small time penalty:
\begin{equation}
r_t \;=\; \underbrace{\mathbbm{1}_{\mathrm{reach}}(t)\,w_{\mathrm{step}}(t)\,r_{\mathrm{score}}(t)}_{\text{sparse subgoal}}
\;+\; \underbrace{\alpha_{\mathrm{dense}}\,r_{\mathrm{dense}}(t)}_{\text{dense tracking}}
\;-\; \underbrace{c_{\mathrm{time}}}_{\text{time}},
\label{eq:reward_full}
\end{equation}

\qheading{Subgoal Indicator $\mathbbm{1}_{\mathrm{reach}}(t)$}
At each step, the active subgoal $g_k$ is considered \emph{instantaneously reached} when all tracking errors fall below their tolerances simultaneously: $e^{\mathrm{pos}}_{t,k} < \epsilon_{\mathrm{pos}}$, $e^{f}_{t,k} < \epsilon_{\mathrm{tip}}$ for every fingertip $f \in \{\text{thumb},\text{index},\text{middle},\text{ring},\text{pinky}\}$, and $e^{\mathrm{rot}}_{t,k} < \epsilon_{\mathrm{rot}}$.
The indicator $\mathbbm{1}_{\mathrm{reach}}(t)$ fires only when this condition is held for $N_{\mathrm{stay}}$ consecutive frames, where $N_{\mathrm{stay}}$ is resampled after every successful subgoal hit.

\qheading{Step Weighting $w_{\mathrm{step}}(t)$}
After a subgoal $g_k$ is hit, the next subgoal is drawn from the same reference trajectory at index $k' = k + \Delta k$, where the jump $\Delta k \in \mathbb{Z}$ is drawn from a uniform distribution whose range expands over training according to the curriculum (\cref{appx:method:curriculum}); $|\Delta k|$ is the number of reference frames skipped between two consecutive subgoals. The step-weighting factor $w_{\mathrm{step}}$ scales the sparse bonus by this temporal gap so that larger jumps yield proportionally larger rewards and the policy is not biased toward exploiting trivially close subgoals. When the environment performs a cross-trajectory switch (\cref{appx:method:episode}), $w_{\mathrm{step}}$ is set to a much larger fixed value for the reward computation at that step instead. Both values are listed in \cref{tab:reward_params}.

\qheading{Subgoal Score $r_{\mathrm{score}}(t)$}
The score blends per-fingertip and object tracking terms with fixed weights:
\begin{equation}
r_{\mathrm{score}} = \alpha_s \Bigl[\; w^s_{\mathrm{tip}} \!\!\sum_{f\in\mathcal{F}}\!\! \rho_{f}
\;+\; w^s_{\mathrm{obj}}\,(\rho_{\mathrm{pos}} + \rho_{\mathrm{rot}})\;\Bigr],
\end{equation}
where the fingertip set $\mathcal{F}$ consists of thumb, index, middle, ring, and pinky for SharpaWave, with the ring finger omitted for LeapHand. Each $\rho_{\cdot}$ is an exponential kernel applied to the corresponding tracking error---the per-fingertip distance $e^{f}_{t,k}$, the object position error $e^{\mathrm{pos}}_{t,k}$, and the object rotation error $e^{\mathrm{rot}}_{t,k}$ in radians:
\begin{equation}
\rho_f = \exp\!\bigl(-\beta_f\, e^f_{t,k}\bigr),\quad
\rho_{\mathrm{pos}} = \exp\!\bigl(-\beta_{\mathrm{pos}}\, e^{\mathrm{pos}}_{t,k}\bigr),\quad
\rho_{\mathrm{rot}} = \exp\!\bigl(-\beta_{\mathrm{rot}}\, |e^{\mathrm{rot}}_{t,k}|\bigr).
\end{equation}
Outer scale $\alpha_s = 1.5$. The remaining blend weights and decay rates $\beta$ (following \citet{li2025maniptrans}) are listed in \cref{tab:reward_params}.

\qheading{Dense Tracking $r_{\mathrm{dense}}(t)$}
Following~\citet{li2025maniptrans}, a small dense signal shapes early exploration before any subgoal is reached:
\begin{equation}
r_{\mathrm{dense}} \;=\; \sum_{i \in \mathcal{I}} w^d_i\,\rho_i \;+\; (1-\sigma_t)\,\bigl(w^d_{\mathrm{pos}}\,\rho_{\mathrm{pos}} + w^d_{\mathrm{rot}}\,\rho_{\mathrm{rot}}\bigr),
\label{eq:r_dense}
\end{equation}
where $\mathcal{I} = \{\text{thumb, index, middle, ring, pinky, lvl1, lvl2}\}$, with \emph{lvl1} the MCP (root) knuckles and \emph{lvl2} the medial (proximal) knuckles, each averaged across fingers, sharing the same exponential kernel form as the per-finger $\rho_f$. The $(1-\sigma_t)$ factor turns the dense object signal off early in training and ramps it in as the curriculum hardens (\cref{appx:method:curriculum}). Weight values $w^d_i$ and decay rates $\beta$ are listed in \cref{tab:reward_params}; the overall dense-reward scale in \cref{eq:reward_full} is $\alpha_{\mathrm{dense}} = 0.1$.

\begin{table}[!h]
\centering
\captionsetup{font=small}
\caption{\textbf{Reward parameters.} Full set of constants used in the reward function (\cref{eq:reward_full}), grouped by reward component.}
\label{tab:reward_params}
\small
\setlength{\tabcolsep}{5pt}
\begin{minipage}[t]{0.48\linewidth}
\centering
\begin{tabular}{lc}
\toprule
Parameter & Value \\
\midrule
\multicolumn{2}{l}{\textit{Subgoal indicator}} \\
\midrule
$\epsilon_{\mathrm{pos}}$ (object pos.\ tolerance)        & $1$\,cm \\
$\epsilon_{\mathrm{tip}}$ (fingertip tolerance)           & $3$\,cm \\
$\epsilon_{\mathrm{rot}}$ (object rot.\ tolerance)        & $10^{\circ}$ \\
$N_{\mathrm{stay}}$ (dwell duration)                       & $\mathcal{U}\{5,\,15\}$ \\
\midrule
\multicolumn{2}{l}{\textit{Step weighting $w_{\mathrm{step}}$}} \\
\midrule
in-traj                                                    & $|\Delta k| + 5$ \\
cross-traj                                                 & $100$ \\
\midrule
\multicolumn{2}{l}{\textit{Subgoal score $r_{\mathrm{score}}$}} \\
\midrule
$\alpha_s$ (outer scale)                                   & $1.5$ \\
$w^s_{\mathrm{tip}}$ (per-fingertip)                       & $0.5$ \\
$w^s_{\mathrm{obj}}$ (pos / rot)                           & $2.0$ \\
\midrule
\multicolumn{2}{l}{\textit{Time penalty}} \\
\midrule
$c_{\mathrm{time}}$                                        & $0.1$ \\
\bottomrule
\end{tabular}
\end{minipage}\hfill
\begin{minipage}[t]{0.48\linewidth}
\centering
\begin{tabular}{lc}
\toprule
Parameter & Value \\
\midrule
\multicolumn{2}{l}{\textit{Kernel decay rates $\beta$}} \\
\midrule
$\beta_{\mathrm{thumb}}$                                                          & $100$ \\
$\beta_{\mathrm{index}} = \beta_{\mathrm{middle}} = \beta_{\mathrm{ring}} = \beta_{\mathrm{pinky}}$ & $90$ \\
$\beta_{\mathrm{lvl1}}$                                                            & $50$ \\
$\beta_{\mathrm{lvl2}}$                                                            & $40$ \\
$\beta_{\mathrm{pos}}$                                                             & $80$ \\
$\beta_{\mathrm{rot}}$ (rad)                                                       & $3$ \\
\midrule
\multicolumn{2}{l}{\textit{Dense tracking $r_{\mathrm{dense}}$}} \\
\midrule
$w^d_{\mathrm{thumb}}$                                                             & $1.0$ \\
$w^d_{\mathrm{index}} = w^d_{\mathrm{middle}} = w^d_{\mathrm{ring}} = w^d_{\mathrm{pinky}}$ & $0.8$ \\
$w^d_{\mathrm{lvl1}}$                                                              & $0.6$ \\
$w^d_{\mathrm{lvl2}}$                                                              & $0.4$ \\
$w^d_{\mathrm{pos}} = w^d_{\mathrm{rot}}$                                          & $1.5$ \\
$\alpha_{\mathrm{dense}}$ (outer scale)                                            & $0.1$ \\
\bottomrule
\end{tabular}
\end{minipage}
\end{table}

\qheading{Time Penalty}
A constant per-step cost $c_{\mathrm{time}} = 0.1$ pressures the policy to complete subgoals quickly and prevents it from lingering in locally stable but unproductive configurations.

\subsection{Curriculum Schedule}
\label{appx:method:curriculum}

Following the curriculum design of \citet{li2025maniptrans}, we progressively harden four difficulty knobs over training. Three of them are jointly controlled by a scalar progress factor $\sigma_t$ that decays from $1$ to $\sigma_{\min} = 0.7$, and simulator gravity follows its own schedule. The four knobs are:
\begin{itemize}[leftmargin=*,noitemsep,nolistsep,topsep=0pt,partopsep=0pt]
  \item \textbf{Inter-subgoal step size} (driven by $\sigma_t$). After each successful subgoal hit (\cref{appx:method:episode}), the next subgoal is drawn $|\Delta k|$ reference frames ahead, with the upper bound on $|\Delta k|$ growing from $40$ frames ($\sim 0.67$\,s at the $60$\,Hz reference rate) to $\sim 80$ frames ($\sim 1.33$\,s) over training. Closer subgoals are easier to reach, so the policy first learns to reach nearby subgoals and only later is asked to reach farther ones.
  \item \textbf{Action-masking duration} (driven by $\sigma_t$). The freeze duration of the random action mask (\cref{appx:method:mask}) is sampled uniformly from $\{1, \dots, d^{\max}_t\}$, with the upper bound $d^{\max}_t$ growing from $1$ to $10$ frames over training. Shorter freezes are easier to tolerate, so the policy first faces brief freezes and only later is exposed to longer ones.
  \item \textbf{Dense object reward} (driven by $\sigma_t$). The object-tracking term $(\rho_{\mathrm{pos}} + \rho_{\mathrm{rot}})$ in $r_{\mathrm{dense}}$ (\cref{eq:r_dense}) is scaled by $(1-\sigma_t)$, growing from zero early on to its full value late, so the policy first learns to track the fingertips and only later learns to track the object pose.
  \item \textbf{Simulator gravity} (independent schedule). Gravity is annealed linearly from $0$ to $-9.8$\,m/s$^2$ over the first $32$\,K environment frames, so the policy first learns to manipulate under reduced object weight and only later has to support the full deployment load.
\end{itemize}

\qheading{Schedules}
The progress factor $\sigma_t$ decays linearly over an annealing window of $T_\sigma = 25{,}600$ environment frames (counted per env), where $t$ is the per-env frame counter:
\[
\sigma_t = 1 - (1-\sigma_{\min}) \cdot \min\!\bigl(t/T_\sigma,\,1\bigr).
\]
Each $\sigma_t$-driven upper bound expands cubically toward its saturation value $u_{\max}$ from its initial value $u_{\min}$,
\[
u^{\max}_t = u_{\min} + \frac{u_{\max} - u_{\min}}{1 - \sigma_{\min}^3}\,\bigl(1 - \sigma_t^3\bigr),
\]
instantiated with $(u_{\min}, u_{\max}) = (40, 80)$ for the inter-subgoal step bound $k^{\max}_t$ and $(1, 10)$ for the action-mask duration bound $d^{\max}_t$.

\subsection{Reset Conditions}
\label{appx:method:episode}

\qheading{Goal Reset}
Whenever a subgoal $g_k$ is hit, the next reference trajectory $\tau_{\mathrm{next}}$ and subgoal index $k_{\mathrm{next}}$ are sampled as
\begin{equation}
(\tau_{\mathrm{next}},\, k_{\mathrm{next}}) \;=\;
\begin{cases}
(\tau,\; k + \Delta k) & \text{with probability } 0.9, \\[2pt]
(\tau' \sim \mathrm{Unif}(\mathcal{D}_o),\; k) & \text{with probability } 0.1 \quad\text{(\emph{cross-trajectory switch})},
\end{cases}
\label{eq:goal_reset}
\end{equation}
where $\tau$ is the current trajectory, $\mathcal{D}_o$ is the reference set for the current object $o$, and $\Delta k$ is drawn from a uniform distribution whose range expands over training according to the curriculum (\cref{appx:method:curriculum}). The next subgoal is then read as the $k_{\mathrm{next}}$-th frame of $\tau_{\mathrm{next}}$. A cross-trajectory switch swaps the trajectory while preserving the frame index. Its purpose is to train the policy to handle transitions between different demonstrations rather than overfitting to any single one. This generality is required at deployment, where the operator's motion will not stay within any recorded trajectory.

\qheading{Episode Reset}
An episode ends if (i) any joint velocity or object linear or angular velocity exceeds its safety bound, (ii) the object position error exceeds $15$\,cm, or (iii) the policy accumulates too many out-of-tolerance frames before reaching the next subgoal:
\[
n_{\mathrm{fail}} \!\leftarrow\!
\begin{cases}
n_{\mathrm{fail}} + 1, & \text{frame out of tolerance},\\
0, & \text{subgoal hit},
\end{cases}
\qquad
\text{terminate if } n_{\mathrm{fail}} > n^{\max}_{\mathrm{fail}} = \eta_{\mathrm{fail}}\,|\Delta k|,
\]
where ``out of tolerance'' means \emph{any} of the five fingertip, object position, or object rotation errors is above threshold, and $\eta_{\mathrm{fail}}$ is the failure-tolerance scale, meaning the policy is allowed $\eta_{\mathrm{fail}}$ out-of-tolerance frames per unit of subgoal step before the episode terminates. We use $\eta_{\mathrm{fail}} = 1.5$. Immediately after a cross-trajectory switch, $n^{\max}_{\mathrm{fail}}$ is overridden by a flat $300$ frames to accommodate the longer regrasp transition. On termination, the environment is re-initialized via \emph{reference state initialization} (RSI)~\citep{li2025maniptrans}: a random frame in the first $90\%$ of the assigned trajectory sets hand DoFs from the retargeted $\mbold{q}^*$ and the object pose from the reference.

\subsection{Geometry-Aware Retargeting Details}
\label{appx:method:retargeting}

We define each loss term in \cref{eq:retarget} and specify the two-stage optimization procedure and loss weights below. The main paper provides compact inline definitions of each loss term; here we give the full formulation together with the optimization procedure and hyperparameters.
Let $\mathcal{H}_t$ denote the robot hand model at frame $t$ with joint configuration $\mbold{q}_t$ and wrist pose $(\mbold{p}^w_t, R^w_t)$, and $\mathcal{O}_t$ the manipulated object mesh transformed by the recorded object pose $T^o_t$. Each link of $\mathcal{H}_t$ carries a precomputed surface point cloud and, for self-collision, a set of body-attached collision spheres. All signed distances $\mathrm{sdf}_{\mathcal{O}_t}(\cdot)$ from a point on the hand surface to the object (positive outside) are computed with NVIDIA Kaolin~\citep{kaolin}, which provides a differentiable point-to-mesh SDF used by both $\mathcal{L}_{\mathrm{surf}}^t$ and $\mathcal{L}_{\mathrm{pen}}^t$ below.

\qheading{Vector Alignment $\mathcal{L}_{\mathrm{vec}}^t$}
We use the standard vector-retargeting loss of~\citet{qin2023anyteleop,handa2020dexpilot} without modification: a weighted Huber on per-vector errors between the captured operator hand keypoints (\cref{appx:method:hoi}) and the corresponding robot vectors. We refer readers to the original papers for the exact loss form, keypoint vector set, and per-vector weights.

\qheading{Surface Attraction $\mathcal{L}_{\mathrm{surf}}^t$}
In the second stage we incorporate the object mesh: points on the hand surface that fall within a threshold $\tau_{\mathrm{surf}}$ of the object are pulled onto the surface,
\[
\mathcal{L}_{\mathrm{surf}}^t = \frac{1}{|\mathcal{S}_t|}\sum_{p \in \mathcal{S}_t} \mathrm{ReLU}(\mathrm{sdf}_{\mathcal{O}_t}(p)),
\quad \mathcal{S}_t = \{p : \mathrm{sdf}_{\mathcal{O}_t}(p) < \tau_{\mathrm{surf}}\}.
\]
This glues the contact-side of the hand to the object without forcing non-contacting links onto it.

\qheading{Mesh Interpenetration $\mathcal{L}_{\mathrm{pen}}^t$}
A symmetric penetration penalty pulls any hand surface point that has entered the object back out:
\[
\mathcal{L}_{\mathrm{pen}}^t = \sum_{p \in \mathcal{H}_t} \mathrm{ReLU}\!\bigl(-\mathrm{sdf}_{\mathcal{O}_t}(p)\bigr).
\]
The weights $\lambda_{\mathrm{pen}}$ and $\lambda_{\mathrm{surf}}$ are listed in the optimization paragraph below.

\qheading{Self-Collision $\mathcal{L}_{\mathrm{col}}^t$}
Self-collision is computed between collision spheres on \emph{different} fingers: for every pair of spheres $(i, j)$ with finger indices $f(i) \ne f(j)$, radii $r_i, r_j$ and centers $\mbold{c}_i, \mbold{c}_j$,
\[
\mathcal{L}_{\mathrm{col}}^t = \tfrac{1}{2}\sum_{f(i)\ne f(j)} \mathrm{ReLU}\!\bigl((r_i + r_j) - \|\mbold{c}_i - \mbold{c}_j\|_2\bigr).
\]
Intra-finger sphere pairs are ignored because adjacent links are designed to touch.

\qheading{Trajectory Smoothness $\mathcal{L}_{\mathrm{smooth}}$}
We adopt the temporal smoothness energy of Curobo~\citep{curobo_v2} without modification: a log-cosh penalty on the velocity, a squared $\ell_2$ penalty on the acceleration, and a  squared $\ell_2$ penalty on the jerk of the per-frame joint configuration $\mbold{q}_t$, summed across the trajectory with internal coefficients fixed to the Curobo defaults. 

\qheading{Optimization}
The two stages are run sequentially per trajectory:
\begin{itemize}[leftmargin=*,noitemsep,nolistsep,topsep=0pt,partopsep=0pt]
  \item \textbf{Stage 1 (vector retargeting).} We minimize $\sum_t \mathcal{L}_{\mathrm{vec}}^t$ jointly over all frames $\mbold{q}_{1:T}$ with Adam, using a GPU-batched FK implementation that follows the dex-retargeting design~\citep{qin2023anyteleop}. The learning rate is $10^{-3}$ and we run $6{,}000$ iterations. This gives a strong but contact-blind initialization.
  \item \textbf{Stage 2 (geometry-aware post-optimization).} From the Stage 1 solution we minimize the remaining loss terms of \cref{eq:retarget} with Adam. The learning rate is $3\!\times\!10^{-3}$, we run $160$ iterations, and the gradient norm is clipped to $1.0$. The surface-attraction mask $\mathcal{S}_t$ is built once at the start of Stage 2 with $\tau_{\mathrm{surf}} = 2$\ and frozen for all iterations.
\end{itemize}
We set $\lambda_{\mathrm{surf}}=10$, $\lambda_{\mathrm{pen}}=2$, $\lambda_{\mathrm{col}}=10$, $\lambda_{\mathrm{smooth}}=0.1$ for both SharpaWave and LeapHand. The final per-frame robot configuration $\mbold{q}^*_t$, together with the recorded object pose $T^o_t$, forms the reference motion consumed by RL.

\subsection{Hand--Object Reference Motion Construction}
\label{appx:method:hoi}

For each object, we record $150$ reference trajectories of unscripted hand--object interactions using the NOKOV MoCap system (\cref{appx:setup:mocap}) at $30$\,Hz. Each trajectory lasts $20$\,s ($600$ frames), giving a total of $\sim 50$ minutes of interaction data per object. The interactions span three categories: (1)~in-hand translation, (2)~in-hand rotation, and (3)~free-play combining arbitrary grasps, finger gaiting, and tool-use sequences. This diversity ensures the reference set covers the full range of contact modes the controller may encounter at deployment.

\begin{figure}[!h]
    \centering
    \includegraphics[width=\linewidth]{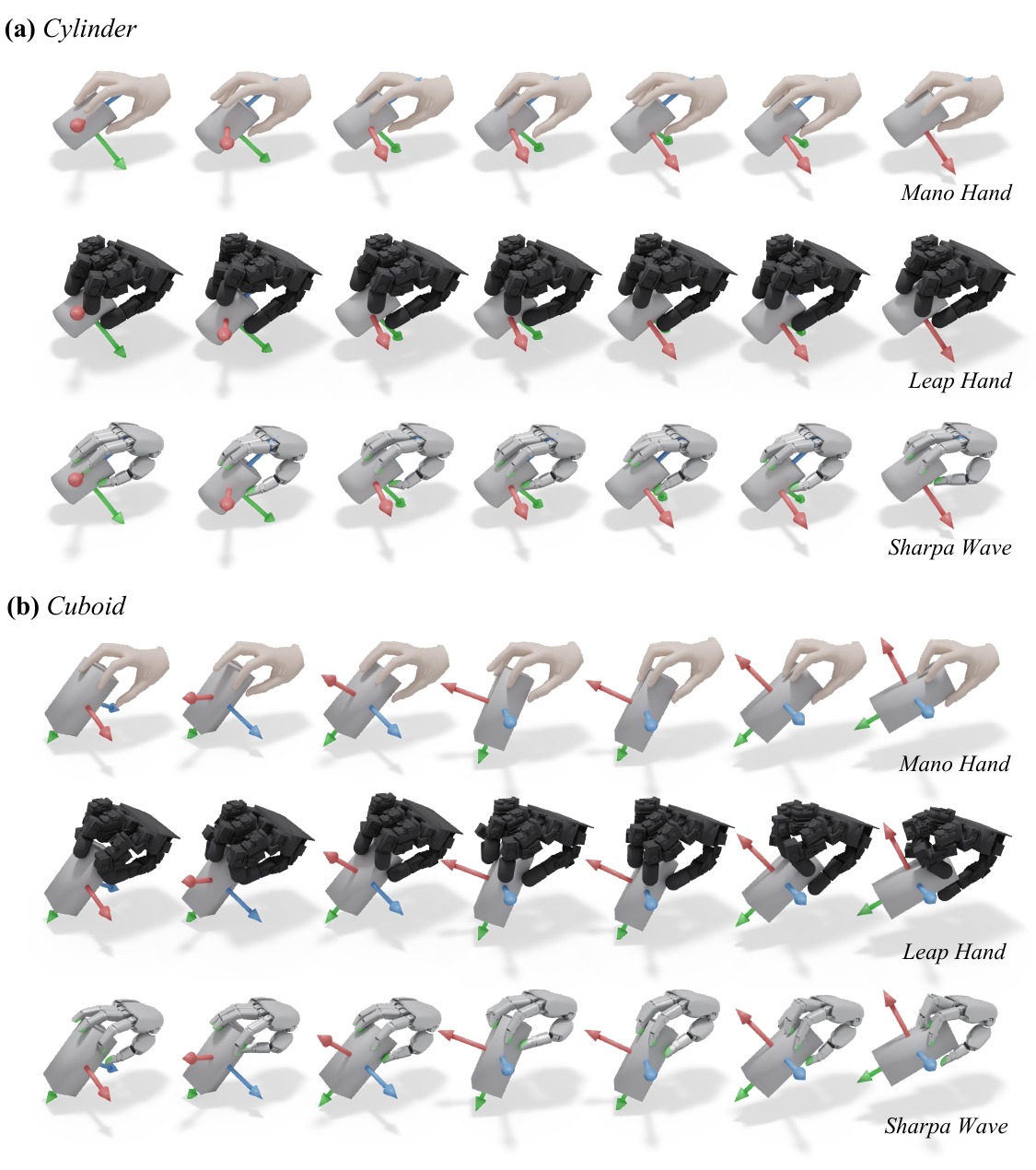}
    \caption{\textbf{Hand--object reference motion visualization.} Retargeted motion clips for (a)~\texttt{Cylinder} and (b)~\texttt{Cuboid}. Each clip shows the source MANO hand and the corresponding retargeted LeapHand and SharpaWave sequences. Coordinate axes indicate the object $6$-D pose.}
    \label{fig:hoi_vis}
\end{figure}

\cref{fig:hoi_vis} visualizes short reference-motion clips for \texttt{Cylinder} and \texttt{Cuboid}, each showing the source MANO hand alongside the retargeted LeapHand and SharpaWave configurations. The retargeted robot hands accurately reproduce the operator's grasp poses and contact transitions across both embodiments, confirming that the two-stage retargeting pipeline (\cref{appx:method:retargeting}) transfers contact-rich interaction structure despite the significant kinematic differences between the human hand and the two robot platforms.

\subsection{Domain Randomization}
\label{appx:method:dr}

We randomize hand and object dynamics, external perturbations, sensing noise, and observation latency to tolerate the sim-to-real gap. Each parameter is sampled independently at the start of every episode and held fixed throughout. The implementation of random force perturbation follows VisualDexterity~\citep{chen2023visual}. The full set of ranges is listed in \cref{tab:dr_ranges}.

\begin{table}[!h]
\centering
\captionsetup{font=small}
\caption{\textbf{Domain randomization ranges.}}
\label{tab:dr_ranges}
\small
\setlength{\tabcolsep}{4pt}
\begin{tabular}{lllc}
\toprule
Group & Parameter & Operation & Range \\
\midrule
\multirow{4}{*}{Hand dynamics}
 & body mass                          & scaling                          & $\mathcal{U}[0.9,\,1.2]$ \\
 & shape friction                     & absolute                          & $\mathcal{U}[1.0,\,4.0]$ \\
 & DoF stiffness $K_p$                & scaling                           & $\mathcal{U}[0.8,\,1.2]$ \\
 & DoF damping $K_d$                  & scaling                           & $\mathcal{U}[0.8,\,1.2]$ \\
\midrule
\multirow{5}{*}{Object physics}
 & body mass                          & scaling                           & $\mathcal{U}[0.5,\,2.0]$ \\
 & surface friction                   & absolute                          & $\mathcal{U}[0.5,\,4.0]$ \\
 & rolling / torsional friction       & absolute                          & $\mathcal{U}[0,\,0.05]$ \\
 & restitution                        & additive                          & $\mathcal{U}[0,\,1.0]$ \\
 & mesh scale                         & scaling                           & $\mathcal{U}[0.95,\,1.05]$ \\
\midrule
\multirow{3}{*}{External force on object}
 & trigger probability per step       & ---                              & $\mathcal{U}[0.01,\,0.25]$ \\
 & force scale                        & constant                        & $1.0$ \\
& exponential decay                  & constant                         & $0.99$ \\
\midrule
\multirow{4}{*}{Sensing noise}
 & joint position $\mbold{q}_t$       & additive Gaussian                & $\sigma = 0.1$ \\
 & fingertip position                 & additive uniform                 & $\pm 5$\,mm \\
 & object position                    & additive uniform                 & $\pm 5$\,mm \\
 & object orientation                   & additive uniform                 & $\pm 2^\circ$ \\
\midrule
\multirow{2}{*}{Observation latency}
 & queue size                         & constant                         & $2$ frames \\
 & queue sampling probability         & constant                         & $0.5$ \\
\midrule
\multirow{2}{*}{Initial state}
 & wrist orientation                  & additive                          & $\pm 30^\circ$ \\
 & reference frame index              & uniform                          & first $90\%$ of clip \\
\bottomrule
\end{tabular}
\end{table}

\subsection{Random Action Masking Details}
\label{appx:method:mask}

Random action masking is the strong action-space regularizer introduced in \cref{sec:method:learning}. It prevents the policy from overfitting to the perfectly synchronized actuation of simulation, which is unrealistic on hardware where actuator compliance, backlash, and PD response vary across DoFs.

\qheading{Mechanism}
At each environment step, with probability $p_{\mathrm{mask}} = 0.15$ \emph{and only when no mask is currently active}, we sample a fresh mask: $n_m = 3$ DoF indices are drawn uniformly without replacement, and a freeze duration $d \sim \mathrm{Uniform}\{1, d^{\max}_t\}$ is drawn (where $d^{\max}_t$ ramps from $1$ to $10$ following the curriculum schedule in \cref{appx:method:curriculum}). For the next $d$ control steps, the executed action $\tilde{\mbold{a}}_t$ on the masked DoFs is overwritten with the previously commanded action, while the unmasked DoFs receive the current policy output:
\[
\tilde{\mbold{a}}_t[j] = \begin{cases} \tilde{\mbold{a}}_{t-1}[j] & j \in \mathcal{M}_t \\ \mbold{a}_t[j] & \text{otherwise}\end{cases}.
\]
After $d$ steps the mask deactivates, and a new mask can be sampled on the next step. Because the mask refreshes asynchronously across the $\sim\!62$\,K parallel environments, training sees a wide spectrum of partially-stale joint commands.

\qheading{Sim-to-Real Effect}
Random action masking effectively augments the training distribution with desynchronized, partially stale joint commands. The policy is therefore forced to recover useful contact configurations even when some joints respond late or not at all, which closely matches the dominant failure modes we observe on the real hardware (motor lag,  backlash, occasional missed commands on the SharpaWave SDK). Empirically, masking is the single most impactful sim-to-real intervention we tested (\cref{tab:ablation_am} in the main paper; further analysis in \cref{appx:results:failure}).

\subsection{RL Training Hyperparameters and Compute}
\label{appx:method:rl}
\label{appx:rl:hparams}   %
\label{appx:rl:compute}   %

\cref{tab:sapg_hparams} provides the full set of network, PPO, and SAPG-specific hyperparameters used to train the controller. All our controllers are trained on $4$ NVIDIA RTX~5090 GPUs running $15{,}600$ parallel environments per GPU ($62{,}400$ environments in total). Each controller is trained for $\sim 10^{10}$ environment steps in a single RL stage, which takes approximately $1$~day on this setup.

\begin{table}[!h]
\centering
\captionsetup{font=small}
\caption{\textbf{Hyperparameters of SAPG.}}
\label{tab:sapg_hparams}
\small
\setlength{\tabcolsep}{8pt}
\begin{tabular}{ll}
\toprule
Hyperparameter & Value \\
\midrule
\multicolumn{2}{l}{\textit{Network}} \\
\midrule
LSTM hidden units                    & $512$ \\
LSTM Layer Normalization             & enabled \\
LSTM sequence length                 & $4$ \\
MLP hidden sizes                     & $[512,\,1024,\,1024,\,512,\,512]$ \\
Activation                           & ELU \\
\midrule
\multicolumn{2}{l}{\textit{PPO}} \\
\midrule
Learning rate                        & $2 \times 10^{-4}$ \\
LR schedule                          & adaptive \\
KL threshold                         & $0.008$ \\
Num opt-epochs                       & $4$ \\
Minibatch size (per GPU)             & $31{,}200$ \\
Horizon length                       & $32$ \\
Discount ($\gamma$)                  & $0.99$ \\
GAE $\lambda$                        & $0.95$ \\
Clip range ($\epsilon$)              & $0.2$ \\
Max grad norm                        & $1.0$ \\
Bounds-loss coef.                    & $10^{-4}$ \\
Parallel envs (per GPU)              & $15{,}600$ \\
\midrule
\multicolumn{2}{l}{\textit{SAPG}} \\
\midrule
Num blocks               & $6$ \\
Entropy Bonus Scale                  & $0.005$ \\
Off-policy ratio                     & $1.0$ \\
Mix ratio                            & $0.5$ \\
\bottomrule
\end{tabular}
\end{table}

\section{Baseline Implementation Details}
\label{appx:baselines}

This section describes how each baseline in \cref{tab:task_results} is implemented and what we change relative to the original release. All baselines are deployed on the same hardware platform and share the same teleoperation interface (\cref{appx:setup:hardware}).

\qheading{DexRT~\citep{handa2020dexpilot,qin2023anyteleop}}
We use the open-source dex-retargeting codebase\footnote{\url{https://github.com/dexsuite/dex-retargeting}} with vector-alignment retargeting. The key parameter is the fingertip scaling factor, set to $1.0$ for SharpaWave and $1.2$ for LeapHand.

\qheading{GeoRT~\citep{yin2025geometric}}
We follow the original paper and official implementation\footnote{\url{https://github.com/facebookresearch/GeoRT}} to train and deploy the neural retargeter. The fingertip workspace data used for training is collected with our own inference glove to match the operator's hand kinematics at deployment.

\qheading{DexGen~\citep{yin2025dexteritygen}}
No official implementation is available for DexGen. We re-implement its foundation dexterity controller following the training and deployment recipe described in the original paper. Because the key intermediate step (the AnyGrasp-to-AnyGrasp RL policy) lacks sufficient detail for faithful reproduction, we substitute it with our co-tracking controller to generate the simulation rollouts, matching the data scale reported in the original paper. All subsequent stages (diffusion-based action prior training and deployment) follow the original design.

\qheading{SimToolReal~\citep{kedia2026simtoolreal}}
SimToolReal is not a teleoperation method but an object-centric sim-to-real tool-use policy; we include it as a strong reference for learned dexterous manipulation. We follow the official implementation\footnote{\url{https://github.com/tylerlum/simtoolreal}} and re-train on our hardware (Franka FR3 $+$ SharpaWave right hand), adapting the simulation workspace and robot embodiment to match our deployment setup. The three tool categories (hammer, brush, screwdriver) and all other training details follow the original paper. We evaluate both the category-specific variant (SimToolReal$^{\dagger}$, one policy per tool) and the all-categories variant (SimToolReal$^{\ddagger}$, a single policy across tools), as reported in \cref{tab:task_results}.

\section{Autonomous Policy Details}
\label{appx:auto}

We adopt the Conv-UNet Diffusion Policy of~\citet{chi2025diffusion} (DDPM noise predictor with $1$-D temporal convolutions). The policy is conditioned on RGB observations from one third-person and one wrist-mounted camera, each encoded by a separate frozen DINOv2 ViT-S/14 encoder. At each step the policy observes the last $T_o = 2$ frames and predicts an action chunk of length $T_p = 16$. Architecture and training hyperparameters are summarised in \cref{tab:dp_hparams}.

\begin{table}[!h]
\centering
\captionsetup{font=small}
\caption{\textbf{Diffusion Policy hyperparameters.}}
\label{tab:dp_hparams}
\small
\setlength{\tabcolsep}{8pt}
\begin{tabular}{ll}
\toprule
Hyperparameter & Value \\
\midrule
Visual encoder              & DINOv2 ViT-S/14, pretrained, frozen \\
Encoder sharing             & Separate encoders per camera \\
Image resolution            & $240\times 320$; random crop $210\times 280$ \\
Observation horizon $T_o$   & $2$ \\
Action horizon $T_p$        & $16$ \\
Conv-UNet channels          & $[512,\,1024,\,2048]$ \\
Diffusion embedding dim     & $128$ \\
Diffusion steps (train)     & $100$ \\
Diffusion steps (inference) & $16$ (DDIM) \\
Optimizer                   & AdamW ($\beta_1\!=\!0.95$, $\beta_2\!=\!0.999$, wd $10^{-6}$) \\
Learning rate               & $10^{-4}$, cosine schedule, $500$-step warmup \\
Batch size                  & $16$ \\
Training epochs             & $300$ \\
\bottomrule
\end{tabular}
\end{table}

\end{document}